\documentclass[10pt, conference, compsocconf]{IEEEtran}
\IEEEoverridecommandlockouts

\usepackage{amsfonts}
\usepackage{mathrsfs}
\usepackage{amssymb}
\usepackage{wrapfig}
\usepackage{amsmath}
\usepackage{accents}
\usepackage{dsfont}
\usepackage{amsbsy}
\usepackage{setspace}
\usepackage{graphicx}
\usepackage{subfig}
\usepackage{url}
\usepackage{color}
\usepackage{epstopdf}
\usepackage{amssymb}
\usepackage{framed}
\usepackage{multirow}
\usepackage{mathtools}  
\usepackage{algorithm}
\usepackage{tabularx}
\usepackage{diagbox}

\begin{document}

\pagestyle{plain}
\setcounter{page}{1}

\newtheorem{theorem}{Theorem}
\newtheorem{lemma}{Lemma}
\newtheorem{claim}{Claim}
\newtheorem{proposition}{Proposition}
\newtheorem{corollary}{Corollary}
\newtheorem{definition}{Definition}
\newtheorem{assumption}{Assumption}
\newtheorem{remark}{Remark}

\newcommand{\cS}{\mathcal{S}}
\newcommand{\vS}{\mathbf{S}}
\newcommand{\vQ}{\mathbf{Q}}
\newcommand{\bE}{\mathds{E}}
\newcommand{\mc}{\mathcal}
\newcommand{\mb}{\mathbf}
\newcommand{\bs}{\boldsymbol}
\newcommand{\ol}{\overline}
\newcommand{\wt}{\widetilde}
\newcommand{\wh}{\widehat}
\newcommand{\id}{\mathds{1}}
\newcommand{\lc}{LCFL}

\def\argmax{\operatornamewithlimits{arg\,max}}
\def\argmin{\operatornamewithlimits{arg\,min}}

\title{On the Low-Complexity of Fair Learning for Combinatorial Multi-Armed Bandit }
\author{Xiaoyi Wu$^{*}$ \mbox{\hspace{0.4cm}} Bo Ji$^{\dag}$ \mbox{\hspace{0.4cm}} Bin Li$^{*}$
\\ $^{*}$Department of Electrical Engineering, Pennsylvania State University, University Park, PA, USA
\\ $^{\dag}$Department of Computer Science, Virginia Tech, Blacksburg, Virginia, USA\thanks{This work has been supported in part by NSF under the grants CNS-2152610, CNS-2152658, and CNS-2312833, the ARO Grant W911NF-24-1-0103, the Commonwealth Cyber Initiative (CCI), and Nokia
Corporation.}
}
\maketitle

\begin{abstract} 
Combinatorial Multi-Armed Bandit with fairness constraints is a framework where multiple arms form a super arm and can be pulled in each round under uncertainty to maximize cumulative rewards while ensuring the minimum average reward required by each arm.  
The existing pessimistic-optimistic algorithm linearly combines virtual queue-lengths (tracking the fairness violations) and Upper Confidence Bound estimates as a weight for each arm and selects a super arm with the maximum total weight. The number of super arms could be exponential in the number of arms in many scenarios. In wireless networks, due to interference constraints the number of super arms can grow exponentially with the number of arms. Evaluating all the feasible super arms to find the one with the maximum total weight can incur extremely high computational complexity in the pessimistic-optimistic algorithm. To tackle this issue, we develop a low-complexity fair learning algorithm based on the so-called pick-and-compare approach that involves randomly picking $M$ feasible super arms to evaluate. By setting $M$ to a constant, the number of comparison steps in the pessimistic-optimistic algorithm can be reduced to a constant, thereby significantly reducing the computational complexity. The theoretical analysis shows that our low-complexity design sacrifices fairness and regret performance only marginally. Finally, we validate our theoretical results through extensive simulations.
\end{abstract}

\section{Introduction}
Combinatorial Multi-Armed Bandit (CMAB) with fairness constraints is a framework where multiple arms form a super
arm and can be pulled in each round to maximize the cumulative rewards and ensure fairness among arms (i.e., guaranteeing the minimum average reward required by each arm). 
One efficient solution for the CMAB with fairness constraints is the pessimistic-optimistic algorithm \cite{li2019combinatorial,liu2021efficient,wu2023joint,wu2024regular,zhou2022kernelized} that involves combining virtual queue-lengths monitoring the fairness violations and Upper Confidence Bound (UCB) estimates balancing the tradeoff between exploitation and exploration in online learning to quantify a weight for each arm. Then, it pulls a super arm with the maximum total weights among all feasible super arms in each round. In \cite{liu2021efficient}, the authors showed that the pessimistic-optimistic algorithm achieves $\mc{\wt{O}}(\sqrt{T})$ regret over $T$ rounds and zero cumulative fairness violation after a certain number of rounds. These bounds are sharp because the regret bound matches the distribution-independent lower bound for the reward regret in CMAB without fairness constraints \cite{BubCes_12}, and zero fairness constraint violation is the smallest possible. Subsequent works (e.g., \cite{guo2023rectified,guo2022online}) further extend the pessimistic-optimistic algorithm to deal with hard fairness constraints. However, the number of super arms in CMAB under fairness constraints could be exponential in the number of arms in many scenarios.
For example, in the application of timely information delivery in wireless networks, multiple sensing sources need to transmit sensing information to one access point (AP) over unreliable wireless channels. 
To keep the completeness of sensing information
collected at AP, not only the system throughput but also fairness
among sensing sources should be considered. This scenario is analogous to a CMAB problem under fairness constraints.
Furthermore, note that transmissions are subject to interference constraints, which ensure that multiple simultaneous transmissions do not degrade the quality of the received signals below acceptable levels \cite{wyglinski2009cognitive}. As a result, only a feasible subset of these sources can transmit data simultaneously. The number of feasible subsets can be exponential in the number of sensing sources depending on the wireless interference \cite{madan2006cross}.
Consequently, in this scenario, the pessimistic-optimistic algorithm must evaluate all feasible subsets (i.e., all feasible super arms in CMAB) to identify the one with the maximum total weight, making it computationally intensive. These inspire us to develop a fair learning algorithm that greatly reduces computational complexity while preserving the performance of the pessimistic-optimistic algorithm.

Prior works on the low-complexity design for CMAB (e.g., \cite{kang2020low,wang2020restless, liu2021low,zhang2024fast,jourdan2021efficient,rejwan2020top}) did not address fairness constraints and thus cannot be applied to our setting. \cite{xu2020combinatorial} considered the constraint on the number of arms that can be pulled simultaneously in each round and did not address the combinatorial constraints on the feasible super arms, which arise particularly in wireless networks. \cite{wang2024online} leveraged a low-complexity index policy, assigning an index to each arm and pulling the super arm with maximum indices to minimize the cumulative
rewards while ensuring fairness in Restless Multi-armed Bandits (RMAB). However, it still requires evaluating all the super arms, which can cause high computational complexity in wireless network applications. In this paper, we build upon the pessimistic-optimistic algorithm and incorporate the powerful low-complexity design approach, pick-and-compare (PC) (e.g., \cite{tassiulas1998linear, yi2008complexity, chaporkar2008stable,li2019low,eryilmaz2005stable}), to develop a low-complexity fair learning algorithm for CMAB with fairness constraints. Specifically, our algorithm uniformly picks $M$ super arms at random and then selects the super arm with the maximum weight among these randomly picked super arms and the one that was selected in the previous round. Note that by setting $M$ to a constant, we can decrease the number of comparison steps in the pessimistic-optimistic algorithm from exponential in the number of arms to a constant, thereby significantly reducing the computational complexity. 
While \cite{kang2020low,park2021learning} also developed a PC-based low-complexity online learning algorithm, they did not specifically address the fairness constraints and their regret analysis requires quantifying the number of rounds selecting the feasible super arm so that its total UCB estimates is close to that of the optimal super arm in each round. However, it is hard to quantify such a number in the presence of weight involving both queue-lengths and UCB estimates as in the pessimistic-optimistic algorithm. As such, we need to address challenges posed by the distinct dynamics of virtual queue-lengths and UCB estimates, as well as their intricate coupling, to rigorously quantify tradeoffs among regret, fairness, and computational complexity.
Our contributions are listed below:

$\bullet$ We develop a low-complexity fair learning algorithm that linearly combines virtual queue-lengths and UCB estimates as a weight for each arm and uses the PC approach to reduce the computational complexity (cf. Algorithm \ref{alg:LCFL}). By setting $M$ to a constant in Algorithm \ref{alg:LCFL}, we can decrease the number of comparison steps in the pessimistic-optimistic algorithm from exponential in the number of arms to a constant, thereby significantly reducing the computational complexity.

$\bullet$ We reveal the dynamics for both virtual queue-lengths and UCB estimates despite the randomness introduced by our algorithm.
Based on this, we derive an upper bound on the cumulative regret over $T$ consecutive rounds under our proposed algorithm (cf. Proposition \ref{prop:regret}).

$\bullet$ We show that the total virtual queue-lengths of the selected super arms under our algorithm is close to that of the maximum total virtual queue-lengths among all feasible super arms with a high probability. Based on this critical insight, we further show that our proposed algorithm achieves zero cumulative fairness violation after a certain number of rounds (cf. Proposition \ref{prop:zero_violation}). 


$\bullet$ We validate our theoretical findings in synthetic simulation and multi-user interactive and panoramic scene delivery application based on the simulation using the head motion trace dataset (cf. Section \ref{sec:simulation}).

\emph{Note on Notation}: We use bold and script font of a variable to denote a vector and a set, respectively. We use $\mb{x}/\mb{y}$ to denote the component-wise division of the vector $\mb{x}$ and $\mb{y}$. We use $\sqrt{\mb{x}}$ to denote the component-wise square root of the vector $\mb{x}$. Let $\|\mb{x}\|_1$ and $\|\mb{x}\|$ denote the $l_1$ and $l_2$ norm of the vector $\mb{x}$, respectively. We use $a\wedge b$ to denote the $\min\{a,b\}$. 
We use $C(n, k)$ to denote the number of ways to choose a subset of $k$ elements from a fixed set of $n$ elements. We use $f(x)=o(g(x))$ to denote $\lim _{x \rightarrow \infty} f(x)/g(x)=0$ and $f(x)=O(g(x))$ to denote $\limsup _{x \rightarrow \infty} f(x)/g(x)<\infty$ for positive functions $f$ and $g$. We use $f(n)=\widetilde{\mathcal{O}}(g(n))$ to denote $f(n)=O(g(n)\log^kn)$ with some $k>0$.


\section{system model}
\label{sec:model}
We consider a combinatorial multi-armed bandit with $N$ arms, allowing for the simultaneous selection of multiple arms in each round. Let $X_n(t)$ be the reward received by arm $n$ in the $t^{th}$ round if it is pulled. Here, we assume that $\{X_n(t)\}_{t\geq0}$ are independently and identically distributed (i.i.d.) Bernoulli random variables with an unknown mean $\mu_n\in(0,1]$. 
Let $\mu_{\min}\triangleq\min_{n}\mu_n>0$. We use $S_n(t)=1$ to denote that arm $n$ is pulled in round $t$, and $S_n(t)=0$ otherwise. Thus, the received reward $R(t)$ in round $t$ can be expressed as $R(t)\triangleq\sum_{n=1}^{N}X_n(t)S_n(t)$. We define $\mb{S}(t)\triangleq(S_n(t))_{n=1}^{N}$  as a \emph{feasible subset}, i.e., 
a subset of arms that can be pulled simultaneously in round $t$. Let $S_{\max}$ be the maximum number of arms that can be pulled simultaneously in each round. The collection of all feasible super arms is represented by $\mc{S}$, which is assumed to be known for selection and unchanging over time. Let $|\mc{S}|$ represent the number of all feasible subsets. Here, we note that $|\mc{S}|$ might be exponential with respect to the number of arms $N$. 

Our goal is to determine a feasible subset in each round that 1) maximizes the \emph{expected cumulative reward} {over consecutive $T$ rounds}, i.e., $\sum_{t=0}^{T-1}\bE[R(t)]$; 2) ensures \emph{fairness} among arms through guaranteeing each arm $n$ to receive at least the reward $\lambda_n>0$ on average, 
\begin{align*}
\liminf_{T\rightarrow\infty}\frac{1}{T}\sum_{t=0}^{T-1}\bE[X_n(t)S_n(t)]\geq\lambda_n, \forall n=1,2,\ldots,N.    
\end{align*}
3) more importantly, requires less \emph{computational complexity}. 


If we have prior knowledge of the reward statistics (i.e., $\bs{\mu}\triangleq(\mu_n)_{n=1}^{N}$), the first two objectives can be achieved through deploying a randomized stationary strategy $\{q^{*}(\mb{S}),\forall\mb{S}\in\mc{S}\}$. Here, $q^{*}(\mb{S})$ represents the probability of selecting a feasible subset $\mb{S}$ and is the solution to the following optimization problem:
\begin{align}
\max_{q(\mb{S})} &\quad \sum_{\mb{S}\in\mc{S}}q(\mb{S})\sum_{n=1}^{N}\mu_nS_n \\
s.t. &\quad \lambda_n+\delta\leq\sum_{\mb{S}\in\mc{S}}q(\mb{S})S_n\mu_n, \forall n=1,2,\ldots,N,
\end{align}
where $\delta>0$ denotes a ``tightness" constant\footnote{In \cite{guo2023rectified}, the authors further developed the algorithm without slackness constant $\delta$. However, the proposed algorithm still preserves the same computational complexity as the pessimistic-optimistic algorithm. }, and $\lambda_n+\delta\leq\mu_n\leq1$ since $S_n\leq1$. If the number of feasible sets is exponential with respect to the number of arms $N$, the complexity of solving the above optimization problem is extremely high. As such, it is critical to design a low-complexity algorithm that can effectively solve this optimization problem. This is even more challenging in the presence of unknown reward statistics. In such a case, the algorithm needs to quickly infer these unknown statistics (also known as (a.k.a.) exploration) while rapidly activating a subset with large total empirical rewards (a.k.a exploitation). We note that maximizing the expected cumulative rewards can be seen as minimizing the cumulative regret over consecutive $T$ rounds. Here, the cumulative regret is defined as the gap between the expected accumulated reward and the optimal expected reward, i.e.,
\begin{align*}
\text{Reg}(T) \triangleq \sum_{t=0}^{T-1}\sum_{n=1}^{N}\mu_n\left(\bE[S^*_n]-\bE\left[S_n(t)\right]\right),
\end{align*}
where $\bE[S^*_n]\triangleq\sum_{\mb{S}\in\mc{S}}q^*(\mb{S})S_n, \forall n$. 

\section{Algorithm Design}
\label{sec:algorithm}

In this section, we first describe the existing pessimistic-optimistic algorithm (e.g., \cite{li2019combinatorial,liu2021efficient,wu2023joint,wu2024regular}) that achieves the first two goals in the presence of unknown reward statistics but requires high computational complexity. Then, we build upon this pessimistic-optimistic algorithm and develop a low-complexity algorithm by using the idea of so-called PC \cite{tassiulas1998linear, yi2008complexity, chaporkar2008stable,li2019low,eryilmaz2005stable} that gradually improves the weight of selected subsets while requiring much less computational complexity.



Recall that the pessimistic-optimistic algorithm combines UCB estimates to minimize the cumulative regret and virtual queues to ensure fairness among arms. Here, the UCB estimates balance exploitation and exploration in online learning with the goal of achieving minimum cumulative regret, while virtual queues are leveraged to monitor the ``reward debt'', thereby tracking cumulative fairness violations. To calculate the UCB estimate, we introduce the following notations. Let $H_n(t)$ denote the number of rounds arm $n$ has been pulled up to round $t$, i.e., $H_n(t)\triangleq\sum_{\tau=0}^{t-1}S_n(\tau)$. We initialize $H_n(0)=0$, reflecting the system starts at $t=0$. The sample mean of the received rewards of arm $n$ until round $t$ is denoted as $\ol{\mu}_n(t)$, i.e., $\ol{\mu}_n(t)\triangleq\left(\sum_{\tau=0}^{t-1}X_n(\tau)S_n(\tau)\right)/H_n(t)$. If arm $n$ has not been pulled yet until round $t$ (i.e., $H_n(t)=0$), we set $\ol{\mu}_n(t)=1$. Let $w_n(t)$ denote the UCB estimate of arm $n$ in round $t$, which is defined below:
\begin{align}
\label{eqn:UCB}
w_n(t)\triangleq \min\left\{\ol{\mu}_n(t)+\sqrt{\frac{3\log t}{2H_n(t)}}, 1\right\},
\end{align}
where $\sqrt{3\log t/(2H_n(t))}$ is the exploration term that measures the uncertainty of the sample mean $\ol{\mu}_n(t)$. If $H_n(t)$ is small, it implies that there has been less exploration of arm $n$, leading to a less accurate estimate of its sample mean.
Consequently, arm $n$ should be assigned a higher priority for pulling. Here, we use the truncated version of the UCB estimate, since the actual reward for each arm is at most $1$. If arm $n$ has not yet been pulled by round $t$ (i.e., $H_n(t)=0$), we assign it the highest pulling priority by setting $w_n(t)=1$.

To ensure fairness among arms, we introduce a virtual queue for each arm to monitor its ``reward debt'' over rounds. To be specific, we let $Q_n(t)$ represent the virtual queue-length of arm $n$ at the beginning of round $t$. The evolution of $Q_n(t)$ is captured below:
\begin{align}
\label{eqn:virtualQ}
Q_n(t+1)=(Q_n(t)+\lambda_n-S_n(t)X_n(t)+\epsilon)^{+},
\end{align}
where $(x)^{+}\triangleq\max\{x,0\}$, $\epsilon\in(0,1)$ is some positive parameter that ensures $\lambda_n+\epsilon<\mu_n\leq1, \forall n$, and its value will be specified later. We initialize $Q_n(0)=0,\forall n$ since the system begins at $t=0$. 

The pessimistic-optimistic algorithm selects a subset with the maximum total weight of all arms within the subset, where the weight of each arm $n$ in round $t$ is defined as $Q_n(t)+\eta w_n(t)$ and $\eta\geq0$ is some control parameter. 
This algorithm is shown to achieve $\widetilde{\mathcal{O}}(\sqrt{T})$ reward regret and zero cumulative constraint violation after a certain number of rounds.  These bounds are sharp because the regret bound matches the distribution-independent lower bound for the reward regret in CMAB problems without constraints \cite{BubCes_12}, and zero constraint violation is the smallest possible. However, the pessimistic-optimistic algorithm requires comparing the weight of all feasible subsets in each round and thus has high computational complexity, since the number of feasible subsets might be exponential with the number of arms.

As such, we aim to develop an algorithm that significantly reduces the number of comparison steps as in the pessimistic-optimistic algorithm while preserving its regret and fairness violation performance. Here, we leverage the idea of the so-called PC algorithm and develop the following Low-Complexity Fair Learning (\lc{}) algorithm that first randomly picks $M$ feasible subsets of arms and then selects the feasible subset with the maximum weight among these randomly picked feasible subsets and the selected subset in the previous round. Here, $M=1,2,\ldots,|\mathcal{S}|$.
Let $\mc{R}_M(t)$ denote the collection of $M$ feasible subsets of arms randomly picked in round $t$.

\begin{algorithm}
\caption{Low-Complexity Fair Learning (\textbf{LCFL}) Algorithm}
\label{alg:LCFL}
\begin{flushleft}
In each round $t$, uniformly pick $M$ feasible subsets of arms at random, denoted as $\mc{R}_M(t)$, and select the subset with the maximum weight among these randomly picked subsets $\mc{R}_M(t)$ and the selected subset $\wh{\mb{S}}(t-1)$ in round $t-1$, i.e., 
\begin{align*}
\wh{\mb{S}}(t)\in\argmax_{\mb{S}\in
\{\mc{R}_M(t), \wh{\mb{S}}(t-1) \}}\sum_{n=1}^{N}\left(Q_n(t)+\eta w_n(t)\right)S_n,
\end{align*}
After pulling arms within the subset $\wh{\mb{S}}(t)$, update the UCB estimates $\mb{w}(t)\triangleq(w_n(t))_{n=1}^{N}$  according to \eqref{eqn:UCB} and the virtual queue-lengths $\mb{Q}(t)\triangleq(Q_n(t))_{n=1}^{N}$ according to \eqref{eqn:virtualQ}.
\end{flushleft}
\end{algorithm}

\section{Performance Analysis}
This section starts by comparing the proposed algorithm with existing state-of-the-art approaches and discussing the challenges in evaluating its performance. Subsequently, we derive an upper bound on the cumulative regret over $T$ consecutive rounds under our proposed algorithm in Proposition \ref{prop:regret} and show it achieves zero cumulative fairness violation after a certain number of rounds in Proposition \ref{prop:zero_violation}.

In \lc{} algorithm, a large $M$ increases the chance of picking the subset with the maximum weight in each round at the cost of requiring more comparison steps. In particular, when $M=|\mc{S}|$, our \lc{} algorithm reduces to the pessimistic-optimistic algorithm. Like the pessimistic-optimistic algorithm, the parameter $\eta$ balances the virtual queue-lengths and the UCB estimates. When $\eta=0$, our \lc{} algorithm is essentially the queue-length-based PC algorithm (see \cite{tassiulas1998linear, yi2008complexity, chaporkar2008stable}) that significantly reduces the computational complexity while preserving the network throughput optimality. When $\eta$ goes to infinity, the \lc{} algorithm aligns with the UCB-based PC algorithm proposed in \cite{kang2020low} that does not address the fairness constraints. 

The regret analysis in \cite{kang2020low} needs to quantify the number of rounds selecting the feasible subset so that its total UCB estimates is close to that of the optimal subset in each round. Such quantity is harder to determine in the presence of fairness constraints, where the weight of each subset includes the virtual queue lengths of arms within the subset that evolve over rounds. Moreover, in the presence of virtual queues, the subset with the maximum total weight changes over rounds, which is different from the fixed optimal subset in \cite{kang2020low}. While these issues were addressed in the regret analysis of the pessimistic-optimistic algorithm, we need to carefully characterize the weight difference between our \lc{} algorithm and the pessimistic-optimistic algorithm to preserve both the regret performance and fairness under our \lc{} algorithm. This is particularly challenging due to the following two facts: 1) even though the ``random pick'' step in our \lc{} algorithm ensures the accurate selection of the subset with the maximum weight\footnote{If there are multiple subsets with the maximum total weight, then we select the subset with the smallest index.} consisting of both virtual queue-lengths and UCB estimates (referred to as ``best picks"), the joint evolution of the virtual queue-lengths and UCB estimates between the ``best picks" requires precise characterization; and 2) the subset with the maximum weight is influenced by our \lc{} algorithm and thus the weight difference between the subset selected by \lc{} algorithm and the subset with the maximum weight is harder to quantify.

To address these challenges, we need to characterize the properties of the duration between ``best picks" and the dynamics of UCB estimates.  The following lemma shows the properties of the time duration between the two ``best picks'' and is similar to that in \cite{chaporkar2008stable} except for the second moment and the expectation of the largest order statistics.



\begin{lemma}
\label{lemma:exp_time_gap}
Denote $\mc{T}_0\triangleq\left\{t_{\tau}\right\}_{\tau=0,1,2, \ldots,\tau'}$ as the rounds when the subsets chosen at random include the one that has the maximum weight, i.e., $\mb{{S}^{\dagger}}(t_{\tau}) \in \mc{R}(t_{\tau})$. Here, $\mb{{S}^{\dagger}}(t)\in\argmax_{\mb{S}\in\mc{S}}\sum_{n=1}^{N}(Q_n(t) +\eta w_n(t))S_n$, $t_0 = 0$, and $\tau'$ denotes the number of rounds that the subsets chosen at random include the one that has the maximum weight within $T$ rounds. Let $\Delta t_\tau\triangleq T\wedge t_{\tau+1}-t_\tau, \forall \tau=0,1,\ldots,\tau'$. Then, $\Delta t_\tau, \tau=0,1,2,\ldots,\tau'$ are independently distributed. Moreover, we have
\begin{enumerate}
\item $\bE\left[\Delta t_\tau\right] \leq 1/\alpha(M)$
\item $\bE\left[(\Delta t_\tau -1)^2 \right] \leq (\alpha^2(M)-3\alpha(M)+2)/\alpha^2(M)$
\item $\bE\left[\max_{\tau\in\{0,1,\ldots,\tau' \}} \Delta t_{\tau} \right] \leq 1 - (1+\log T)/(\log(1-\alpha(M))) $
\end{enumerate}
where $\alpha(M)\triangleq M/|\mc{S}|$ denotes the probability that the subsets chosen at random include the one that has the maximum weight in each round. 
\end{lemma}

\begin{IEEEproof} In round $t_{\tau}$, the subsets chosen at random include the one with the maximum weight, and thus, the ``compare'' step in the \lc{} algorithm ensures that the subset with the maximum weight is selected for the play, i.e., $\mb{\wh{S}}(t_{\tau}) = \mb{{S}^{\dagger}}(t_{\tau})$. Since each ``pick'' step in the LCFL algorithm is uniformly random, the events that subsets chosen at random including the one with the maximum weight can be seen as independent Bernoulli trials with probability $\alpha(M)\triangleq C(|\mc{S}|-1,M-1)/C(|\mc{S}|,M) =M/|\mc{S}|$. Therefore, $\Delta t_\tau, \tau=0,1,\ldots,\tau'$ are independently distributed and follow the same geometric distribution except $\Delta t_{\tau'}$.
The detailed proof of $(1),(2),(3)$ is in Appendix \ref{app:lemma:expect_max_geo}.
\end{IEEEproof}

In order to characterize the weight dynamics of our \lc{} algorithm, we need  \cite[lemma 1]{wu2023joint} that captures the dynamics of the UCB estimate of each arm over rounds, as shown below.
\begin{lemma}
\label{lemma:weight_change}
When $t\geq 1$ and $H_n(t-1)\geq 1,\forall n$, the weight gap between any two continuous rounds can be bounded as 
\begin{align}
w_n(t)-w_n(t-1)\geq & -\frac{1}{H_n(t-1)}-\sqrt{\frac{3\log T}{2H_n(t-1)}} .
\end{align}
When $H_n(t-1)=0$ and $\wh{S}_n(t-1)=0$, $w_n(t)-w_n(t-1)=0$.
When $H_n(t-1)=0$ and $\wh{S}_n(t-1)=1$, $w_n(t)-w_n(t-1)\geq -1$.
\end{lemma}

Based on Lemmas \ref{lemma:exp_time_gap} and \ref{lemma:weight_change}, we are ready to analyze the regret performance of our \lc{} algorithm.
\begin{proposition}
\label{prop:regret}
[Cumulative Regret] Under the \lc{} algorithm with $\epsilon\leq \delta/2$, the cumulative regret $\text{Reg}(T)$ over consecutive $T$ rounds can be bounded from above as follows:
\begin{align}
&\text{Reg}(T)\leq \min \bigg\{ \frac{NT}{\eta\mu_{\min}}  +2\sqrt{6NS_{\max}T\log T} +N\left(1+\frac{5\pi^2}{12}\right)  \nonumber\\
& + \frac{2C_1(M) NT}{\eta} + C_2(M) \frac{N\pi^2}{6}  + \frac{1+\log T}{C_3(M)}N\log \frac{S_{\max}T}{N} \nonumber \\
&  + C_2(M) N
 + \frac{\left(1+\log T\right)\sqrt{6NS_{\max}T\log T}}{2C_3(M)}, S_{\max}\mu_{\max}T
\bigg\},
\end{align}
where $C_1(M)\triangleq (\alpha^2(M)-3\alpha(M)+2)/\alpha^2(M)$, $C_2(M)\triangleq 1/\alpha(M) -1$ and $C_3(M)\triangleq -\log (1-\alpha(M))$.
\end{proposition}
\begin{IEEEproof}
Select the Lyapunov function $L(t)\triangleq\frac{1}{2}\sum_{n=1}^{N}Q_n^2(t)/\mu_n $
and perform drift-plus-penalty analysis on 
\begin{align}
\bE\left[L(t+1)-L(t)\right]+\eta\Delta R(t),
\end{align}
where $\Delta R(t)\triangleq\sum_{n=1}^{N}\bE\left[\mu_nS_n^*(t)-\mu_n\wh{S}_n(t)\right]$ and the cumulative regret $\text{Reg}(T)\triangleq\sum_{t=0}^{T-1}\Delta R(t)$. While this is the same as the regret analysis of the pessimistic-optimistic algorithm, we need to decompose the cumulative regret into two parts carefully. The first part is the cumulative gap between UCB estimates generated by our \lc{} algorithm and the true mean $\sum_{t=0}^{T-1}\sum_{n=1}^{N}\bE\left[\left(w_n(t)-\mu_n\right)\wh{S}_n(t)\right]$ and its analysis is similar to that of the pessimistic-optimistic algorithm.  The major difference lies in the second part that captures the gap between the max weight type algorithm for the combined virtual queue-length and true mean and our \lc{} algorithm $$\sum_{t=0}^{T-1}\bE\left[\sum_{n=1}^{N}\left(Q_n(t)+\eta\mu_n\right)\wt{S}_n(t)-\sum_{n=1}^{N}\left(Q_n(t)+\eta w_n(t)\right)\wh{S}_n(t)\right]$$ where $\wt{\mb{S}}(t)\in\argmax_{\mb{S}\in\mc{S}}\sum_{n=1}^{N}\left(Q_n(t)+\eta\mu_n\right)S_n.$ To analyze the second part, we first quantify the weight gap when $t \in \mc{T}_0$ (i.e., ``best pick") and then leverage Lemma \ref{lemma:weight_change} to get sharper dynamics of the UCB estimates. Please see details in Appendix \ref{App:proof:regret}.
\end{IEEEproof}


\begin{remark} 
From the derived regret upper bound in Proposition \ref{prop:regret}, we have the following observations: 1) As $M$ increases, $C_1(M)$ and $C_2(M)$ decreases while $C_3(M)$ increases, implying that the regret performance improves at the cost of increasing the computational complexity. Moreover, the regret performance improves more if we increase $M$ values when $M$ is small, as demonstrated via simulations. 2) When $\alpha(M)$ is close to one (i.e., the number of subsets picked at random is close to $|\mc{S}|$), $C_1(M)$ and $C_2(M)$ are close to zero and $C_3(M)$ approaches to infinity. In this case, our \lc{} algorithm is essentially the pessimistic-optimistic algorithm, and our derived regret bound matches that of the pessimistic-optimistic algorithm. 
3) As in the regret performance of the pessimistic-optimistic algorithm, a larger $\eta$ will yield less cumulative regret, which is consistent with our intuition since a large $\eta$ indicates that our \lc{} algorithm puts more emphasis on the UCB estimates and thus yields less cumulative regret.
\end{remark}
To characterize the cumulative fairness violation performance, we need first to quantify the closeness of total queue-lengths of the feasible subset selected by our \lc{} algorithm to the maximum total queue-lengths of all feasible subsets in each round, as revealed in the following lemma.
\begin{lemma}
\label{lemma:large_prob_weight_gap}
The total virtual queue lengths of the super arm picked by the LCFL algorithm are close to that with the maximum total virtual queue lengths in the following sense:
\begin{align}
\Pr\left\{  \sum_{n=1}^{N}Q_n(t){S}^{\ddagger}_n(t)-\sum_{n=1}^{N}Q_n(t)\wh{S}_n(t) \leq B_1 \right\}  \geq \gamma(M,D), \forall t
\end{align}
where $B_1\triangleq 2ND(1+\eta)$, $\mb{S}^{\ddagger}(t)\in\argmax_{\mb{S}\in \mc{S}}\sum_{n=1}^NQ_n(t)S_{n}$, $\gamma(M,D)\triangleq 1-(1-M/|\mc{S}|)^{D}\geq 1/(\delta+1)$, and $D$ is some positive constant. 
\end{lemma}
\begin{IEEEproof}
The detailed proof is in Appendix \ref{app_lemma:large_prob_weight_gap}.
\end{IEEEproof}

Building upon lemma \ref{lemma:large_prob_weight_gap} and following similar steps in \cite[Lemma 11]{liu2021efficient}, we analyze the cumulative fairness violation performance under our \lc{} algorithm, as shown below. The detailed proof is in Appendix \ref{App:proof:zero_violation}. In the rest of the paper, we omit the arguments of $\gamma(M,D)$ for simplicity.
\begin{proposition} 
\label{prop:zero_violation}
[Zero Cumulative Fairness Violations] Under the \lc{} algorithm, if $\epsilon \leq  (\gamma\delta+\gamma-1)/2  $, there exists a zero violation point $t_0\triangleq g_0(\eta)/\epsilon=O(\eta/\epsilon)$ after which the zero cumulative fairness violation is achieved for any round $t\geq t_0$, i.e., 
\begin{align*}
\sum_{n=1}^{N}\left(\sum_{\tau=0}^{t-1}(\lambda_n-\bE[\wh{S}_n(\tau)X_n(\tau)])\right)^{+}=0, \quad\forall t\geq t_0,
\end{align*}
where $g_0(\eta)\triangleq\sqrt{N}\left((\log(v_0(\gamma)+1))/\theta(\gamma)+\zeta+U(\eta,\gamma) \right)=O(\eta)$, $\zeta\triangleq N/\mu_{\min}$, $ U(\eta,\gamma)\triangleq 2N/(\mu_{\min}(\gamma\delta+\gamma-1))+8 ND(1+\eta)/(\gamma\delta+\gamma-1) $, $\theta(\gamma) \triangleq 3(\gamma\delta+\gamma-1) /(12\zeta^2+ (\gamma\delta+\gamma-1) \zeta)$, and $v_0(\gamma)\triangleq 8/((\gamma\delta+\gamma-1) \theta(\gamma)$.
\end{proposition}


\begin{remark}
From Proposition \ref{prop:zero_violation}, we have the following observations: 1) As $M$ increases (i.e., increasing the computational complexity), $\gamma$ increases, allowing a larger $\epsilon$ and thus achieving zero violation point earlier. Moreover, when $M=|\mc{S}|$ (i.e., our \lc{} algorithm reduces to the pessimistic-optimistic algorithm), our derived zero violation point matches that of the pessimistic-optimistic algorithm. 2) As in cumulative fairness violation performance of the pessimistic-optimistic algorithm, the zero-violation point is inversely proportional to parameter $\epsilon $ used in the virtual queue-length evolution (cf. \eqref{eqn:virtualQ}). This relationship is logically consistent, as a higher $\epsilon$ leads to greater virtual queue-lengths, compelling the \lc{} algorithm to prioritize arms with larger queue-lengths.  Consequently, the algorithm reaches the zero cumulative fairness violation point more rapidly. Additionally, according to Proposition \ref{prop:zero_violation}, a higher $\eta$ parameter delays the zero violation point, aligning with our intuition. Specifically, increasing $\eta$ emphasizes the UCB estimates, causing the \lc{} algorithm to select arms with higher UCB estimates more frequently, thereby delaying the zero cumulative fairness violation point.

\end{remark}

\begin{remark} 
Table \ref{table:tradeoff} outlines three sets of parameters $(\eta,\epsilon, M)$, detailing their effects on cumulative fairness violations, regret, and computational complexity. In particular, we set $\epsilon=O(1/\sqrt[3]{T})$ to ensure that $\epsilon<(\gamma\delta+\gamma-1)/2$ for a large time horizon $T$ and provide three different sets of $\eta$ and $M$ values to reveal the tradeoff among cumulative fairness violations, regret, and computational complexity. 
According to \cite[remark 4]{wu2024regular}, if $\eta=o(\epsilon t)$, i.e., $\eta/\epsilon=o(t)$, then, zero violation point is sublinear with respect to the time horizon $T$ (i.e., $t_0=o(T)$), guaranteeing  
zero cumulative fairness violations. Next, we analyze the cumulative regret and computational complexity with different $\eta$ and $M$ values and how they reduce to the queue-length-based PC algorithm and the pessimistic-optimistic algorithm:

\begin{itemize}
\item [(1)] When $\eta=(\sqrt{T})$, we can see from the second row of Table \ref{table:tradeoff} that as the number of randomly picked subsets $M$ increases (i.e., the computational complexity increases), the regret performance improves and the zero violation point becomes smaller (i.e., it achieves the zero cumulative fairness violation earlier). 
\item [(2)] When $M=|\mc{S}|$ (i.e., all feasible subsets are picked in each round), our \lc{} algorithm is essentially the pessimistic-optimistic algorithm (e.g., \cite{liu2021efficient,li2019combinatorial,wu2023joint,wu2024regular}). Our analysis is tight in the sense that (from the third row of Table \ref{table:tradeoff}) the zero violation point is sublinear and the regret is $\wt{\mc{O}}(\sqrt{T})$, which is similar to that derived in \cite{liu2021efficient}.
\item[(3)] When $\eta=0$, we can see from the last row of Table \ref{table:tradeoff} that zero cumulative fairness violations will be achieved after a finite number of rounds. However, it yields linear regret since the algorithm does not take the UCB estimates into account and thus cannot efficiently learn the reward statistics. 
\end{itemize}

\end{remark}
\begin{table}[h!]\footnotesize
\renewcommand\arraystretch{1.3}
\centering
\scalebox{0.8}{
\begin{tabular}{|l|c|c|c|}
\hline
\diagbox{($\eta$, $\epsilon$, $M$)} {Performance}& Zero Violation Point & Regret & Complexity \\
\hline
$(O(\sqrt{T}),O(1/\sqrt[3]{T}),M)$ &$O((|\mc{S}|\sqrt[6]{T^5})/M)$ & $\wt{\mc{O}}((|\mc{S}|\sqrt{T})/M) $& $M$ \\
\hline
$(O(\sqrt{T}),O(1/\sqrt[3]{T}),|\mc{S}|)$ &$O(\sqrt[6]{T^5})$ & $\wt{\mc{O}}(\sqrt{T}) $& $|\mc{S}|$ \\
\hline 
$(\eta=0, O(1/\sqrt[3]{T}),M)$ &$O(1)$ & $\wt{\mc{O}}(T) $& $M$ \\
\hline
\end{tabular}
}
\caption{Performance tradeoff}
\label{table:tradeoff}
\end{table}

\section{Simulations}
\label{sec:simulation}
In this section, we first perform synthetic simulations to validate the performance of our \lc{} algorithm. Then, we illustrate the efficiency of our proposed \lc{} algorithm in a multi-user interactive and panoramic scene delivery application based on the real motion trace dataset.

\subsection{Synthetic Simulations}

\label{sec:simulation_syn}
\begin{figure*}[!htbp]
\vspace{-0.25in}
\centering 
\subfloat[Average Reward]{
\label{fig:sim:eta_frac}
\includegraphics[scale=0.3]{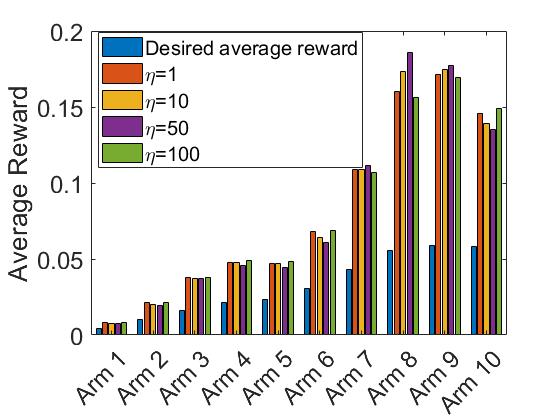}
\hspace{-0.28in}
} 
\subfloat[Cumulative Fairness Violations]{ \label{fig:sim:eta_violation}
\includegraphics[scale=0.3]{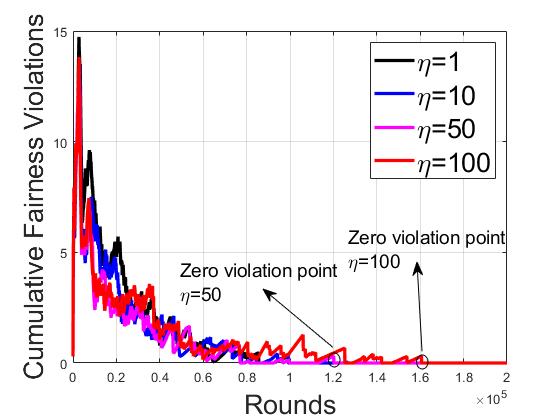}
\hspace{-0.28in}
} 
\subfloat[Cumulative Regret]{ \label{fig:sim:eta_regret}
\includegraphics[scale=0.3]{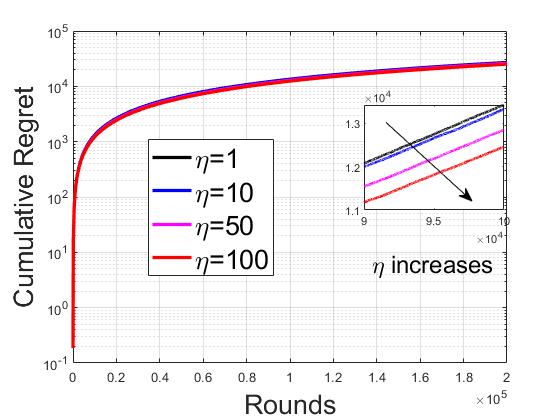}
\hspace{-0.28in}
} 
\caption{Synthetic simulations: impact of parameter $\eta$.}
\label{fig:sim:eta}
\end{figure*}

\begin{figure*}[!htbp]
\vspace{-0.2in}
\centering 
\subfloat[Average Reward]{
\label{fig:sim:num_frac}
\includegraphics[scale=0.3]{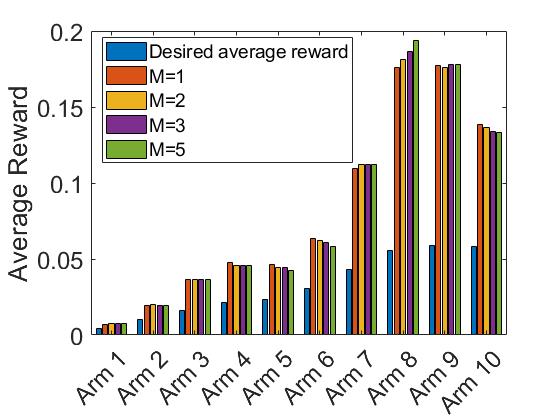}
\hspace{-0.28in}
} 
\subfloat[Cumulative Fairness Violations]{ \label{fig:sim:num_violation}
\includegraphics[scale=0.3]{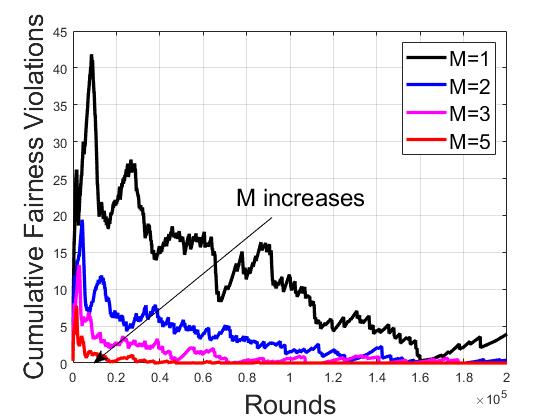}
\hspace{-0.28in}
} 
\subfloat[Cumulative Regret]{ \label{fig:sim:num_regret}
\includegraphics[scale=0.3]{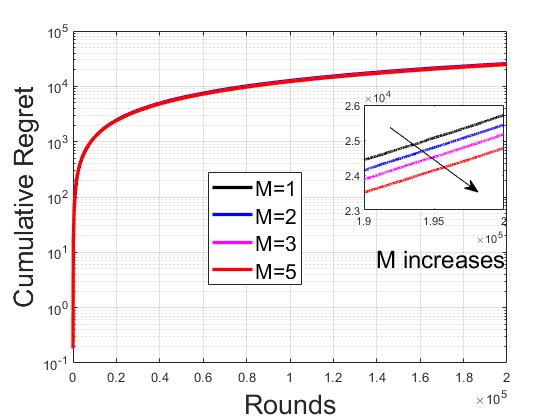}
\hspace{-0.28in}
} 
\caption{Synthetic simulations: impact of parameter $M$.}
\label{fig:sim:num}
\end{figure*}
\begin{figure*}[!htbp]
\vspace{-0.2in}
\centering 
\subfloat[Average Reward]{
\label{fig:sim:trace_eta_frac}
\includegraphics[scale=0.3]{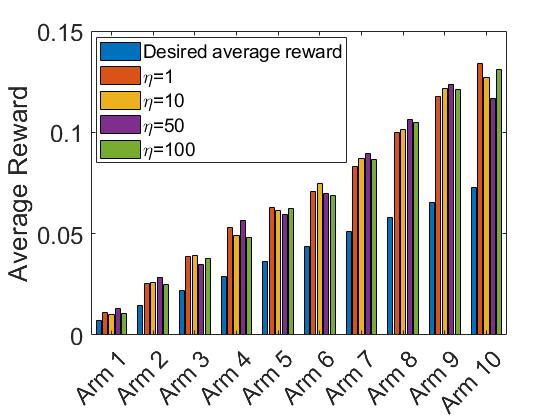}
\hspace{-0.28in}
} 
\subfloat[Cumulative Fairness Violations]{ \label{fig:sim:trace_eta_violation}
\includegraphics[scale=0.3]{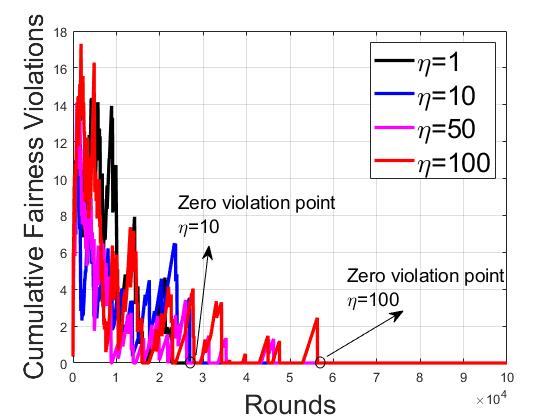}
\hspace{-0.28in}
} 
\subfloat[Cumulative Regret]{ \label{fig:sim:trace_eta_regret}
\includegraphics[scale=0.3]{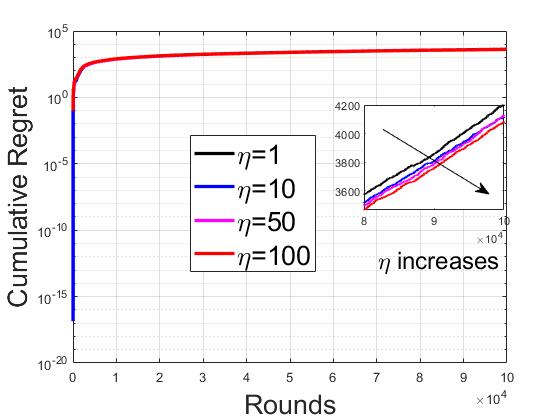}
\hspace{-0.28in}
} 
\caption{Multi-user panoramic scene delivery: impact of parameter $\eta$.}
\label{fig:sim:trace_eta}
\end{figure*}

\begin{figure*}[!htbp]
\vspace{-0.2in}
\centering 
\subfloat[Average Reward]{
\label{fig:sim:trace_num_frac}
\includegraphics[scale=0.3]{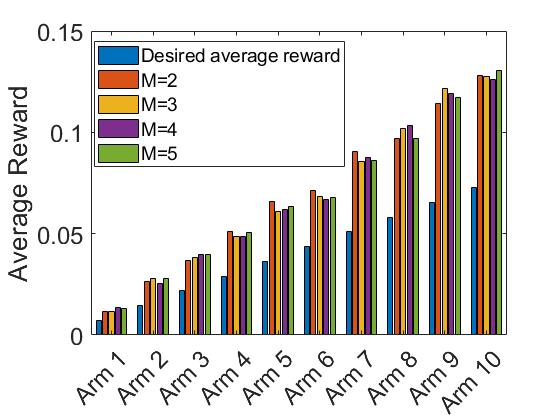}
\hspace{-0.28in}
} 
\subfloat[Cumulative Fairness Violations]{ \label{fig:sim:trace_num_violation}
\includegraphics[scale=0.3]{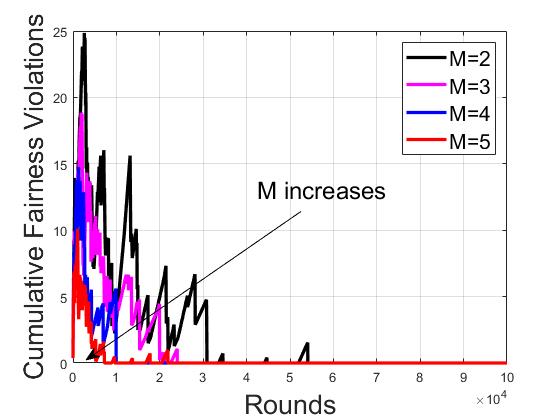}
\hspace{-0.28in}
} 
\subfloat[Cumulative Regret]{ \label{fig:sim:trace_num_regret}
\includegraphics[scale=0.3]{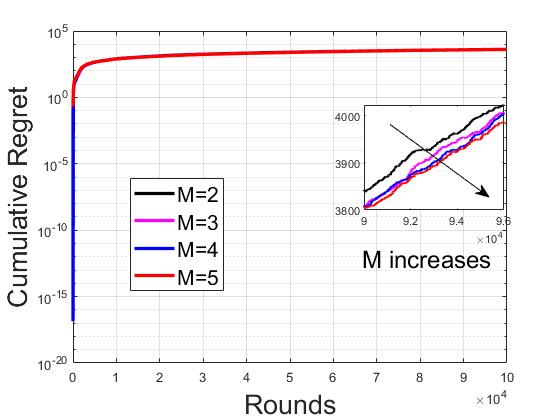}
\hspace{-0.28in}
} 
\caption{Multi-user panoramic scene delivery: impact of $M$.}
\label{fig:sim:trace_num}
\end{figure*}

\label{sim:synthetic}
We consider $N=10$ arms, where at most one arm can be pulled in each round. The mean reward vector is $\bs{\mu}=$(0.6, 0.68, 0.75, 0.73, 0.65, 0.7, 0.85, 0.95, 0.9, 0.8); the minimum required average reward vector is $\bs{\lambda}=0.4\times$(0.6, 1.36, 2.25, 2.92, 3.25, 4.2, 5.95, 7.6, 8.1, 8)/55; and $\epsilon$ is set to $10^{-5}$.

First, we investigate the impact of parameter $\eta$ on the system performance by fixing $M=3$. Fig. \ref{fig:sim:eta} illustrates the performance of our \lc{} algorithm with different $\eta$ values. Fig. \ref{fig:sim:eta_frac} shows the average reward obtained by each arm in our \lc{} algorithm is greater than its minimum required average reward, which indicates that our algorithm satisfies the long-term fairness constraints. Moreover, we can observe from Fig. \ref{fig:sim:eta_violation} that the system achieves zero violation slower with a larger $\eta$. Small $\eta$ (e.g., $\eta=1$ or $\eta=10$) does not have a significant impact on the performance of cumulative fairness violation. 
Compared with the pessimistic-optimistic algorithm, the cumulative fairness violation performance fluctuates more due to the randomness introduced by our \lc{} algorithm. 
A larger $\eta$ postpones the zero violation point but improves the performance of the cumulative regret, as shown in Fig. \ref{fig:sim:eta_regret}. It matches our intuition that a larger $\eta$ represents more emphasis on the UCB estimates, resulting in more frequently pulling the arm with a higher UCB estimate.


Second, we demonstrate how the number of randomly picked feasible subsets $M$ in each round affects the algorithm's performance when $\eta=100$ in Fig. \ref{fig:sim:num}. From Fig. \ref{fig:sim:num_frac}, we can observe that all the arms satisfy the minimum average reward requirement under different $M$ values, achieving the long-term fairness guarantees. We observe from Figs. \ref{fig:sim:num_violation} and \ref{fig:sim:num_regret} that as $M$ increases, the cumulative fairness violation reaches zero earlier, and the regret performance improves. However, this is at the cost of increasing the computational complexity. This is also revealed from our theoretical results (cf. Propositions \ref{prop:zero_violation} and \ref{prop:regret}). In addition, we can see that both cumulative fairness violation and regret performance improve more prominently from $M=2$ to $M=3$ than from $M=1$ to $M=2$. 

\subsection{Motion Trace-based Multi-user Interactive and Panoramic Scene Delivery Simulations}
\label{sec:simulation_trace}

In this subsection, we illustrate the efficiency of our proposed \lc{} algorithm in a multi-user interactive and panoramic scene delivery application based on real motion trace dataset \cite{bao2016shooting}. In our simulations, we consider $N=10$ users. The simulation setting is similar to \cite{wu2024regular}. 
Regarding the fairness constraint, the minimum required average reward vector is $\bs{\lambda}=0.4\times$(0.5, 1.36, 2.25, 2.92, 3.25, 4.2, 5.95, 7.6, 8.1, 8)/55. Parameter $\epsilon$ is set to $10^{-3}$. Fig. \ref{fig:sim:trace_eta} and Fig. \ref{fig:sim:trace_num} illustrate the impact of parameter $\eta$ and $M$ on the cumulative fairness and cumulative regret, respectively. The observations are akin to those presented in Fig. \ref{fig:sim:eta} and Fig. \ref{fig:sim:num} through synthetic simulations.

\section{Conclusion}
In this paper, we studied the low-complexity online learning algorithm design for the CMAB problem with fairness constraints with the goal of minimizing the cumulative regret and guaranteeing fairness among arms. While the existing pessimistic-optimistic algorithm achieves the desired regret and fairness violation performance, it incurs potentially high computational complexity depending on the number of feasible subsets. We developed a low-complexity implementation of the pessimistic-optimistic algorithm based on the so-called PC approach that gradually improves the quality of the weight of the selected subsets. However, unlike the traditional queue-length-based PC algorithm, we need to carefully quantify the dynamics of virtual queue lengths and UCB estimates to obtain the tradeoff among regret, fairness violation, and computational complexity. Simulations were provided to demonstrate the efficiency of our proposed algorithm.

\appendices
\section{Proof of Lemma \ref{lemma:exp_time_gap}}
\label{app:lemma:expect_max_geo}

(1) The first moment for $\Delta t_{\tau},\forall \tau \in \{0,1,\ldots,\tau^{\prime}-1\}$ is
\begin{align}
\label{exp_time_gap_first_moment}
&\bE\left[ \Delta t_{\tau}\right]= \sum_{k=1}^{\infty} k \cdot(1-\alpha(M) )^{k-1} \alpha(M) \nonumber\\
&= \alpha(M)\left(\sum_{k=1}^{\infty}(1-\alpha(M))^{k-1}+\sum_{k=2}^{\infty}(1-\alpha(M))^{k-1}+\cdots \right) \nonumber\\
 & = \alpha(M) \left((1 / \alpha(M))+(1-\alpha(M)) / \alpha(M)+\cdots\right) \nonumber \\
 & = 1+(1-\alpha(M))+(1-\alpha(M))^2+\cdots \nonumber \\
& =  1 /\alpha(M).
\end{align}
When $\tau=\tau^{\prime}$,
\begin{align}
\bE\left[\Delta t_{\tau}\right]=\bE\left[T-t_{\tau}\right] \leq \bE\left[t_{\tau+1}-t_{\tau}\right] =1/\alpha(M).
\end{align}
(2) The second moment for $\Delta t_{\tau}, \forall \tau \in \{0,1,\ldots,\tau^{\prime}-1\}$ is 
\begin{align}
\label{exp_time_gap_second_moment}
&\bE\left[(\Delta t_{\tau})^2\right]  =\sum_{k=1}^{\infty} k^2 (1-\alpha(M))^{k-1} \alpha(M) \nonumber \\
& =\sum_{k=1}^{\infty}(k-1+1)^2 (1-\alpha(M))^{k-1} \alpha(M) \nonumber \\
& =\sum_{k=1}^{\infty}(k-1)^2 (1-\alpha(M))^{k-1} \alpha(M) \nonumber \\
& +\sum_{k=1}^{\infty} 2(k-1) (1-\alpha(M))^{k-1} \alpha(M) \nonumber \\
& +\sum_{k=1}^{\infty} (1-\alpha(M))^{k-1} \alpha(M) \nonumber \\
& =\sum_{j=1}^{\infty} j^2 (1-\alpha(M))^j \alpha(M) \nonumber \\
& +2 \sum_{j=1}^{\infty} j(1-\alpha(M))^j \alpha(M)+1 \nonumber \\
& =(1-\alpha(M))\bE\left[(\Delta t_{\tau})^2\right]+2 (1-\alpha(M)) \bE[\Delta t_{\tau}]+1.
\end{align}

After inserting \eqref{exp_time_gap_first_moment} into \eqref{exp_time_gap_second_moment}, we can have
$$
\bE\left[(\Delta t_{\tau})^2\right] = (2-\alpha(M))/\alpha^2(M),
$$
and 
\begin{align}
&\bE\left[ (\Delta t_{\tau}-1)^2 \right] \nonumber\\
=& \bE\left[(\Delta t_{\tau})^2\right] -2 \bE\left[\Delta t_{\tau}\right]+1\nonumber\\
 = & (\alpha^2(M)-3\alpha(M)+2)/\alpha^2(M).
\end{align}
When $\tau=\tau^{\prime}$,
\begin{align}
\bE\left[ (\Delta t_{\tau}-1)^2 \right] & =    \bE\left[ (T-t_{\tau}-1)^2 \right] \nonumber\\
& \leq \bE\left[ (t_{\tau+1}-t_{\tau}-1)^2 \right] \nonumber \\
 &= (\alpha^2(M)-3\alpha(M)+2)/\alpha^2(M).
\end{align}
(3) Since $\tau' \leq T-1$, we have
$$
\bE\left[\max_{\tau\in\{0,1,\ldots,\tau' \}} \Delta t_{\tau} \right] \leq \bE\left[\max_{\tau\in\{0,1,\ldots,T-1 \}} \Delta t_{\tau} \right] .
$$

Let $q=1-\alpha(M)$. We have $\Pr(\Delta t_{\tau} \leq k)=1-q^k$ to get
\begin{align}
\bE\left[\max_{\tau\in\{0,1,\ldots,T-1 \}} \Delta t_{\tau} \right] & \stackrel{(a)}{=}\sum_{k=0}^{\infty}\Pr\left(\max_{\tau\in\{0,1,\ldots,T-1 \}} \Delta t_{\tau} \geq k \right) \nonumber \\
& \stackrel{(b)}{=}\sum_{k=0}^{\infty}(1- \prod_{\tau=0}^{T-1} \Pr\left(\Delta t_{\tau} \leq k\right)) \nonumber \\
& =\sum_{k=0}^{\infty}\left(1-\left(1-q^k\right)^T\right) \nonumber \\
&\stackrel{(c)}{\leq}  1+\int_0^{\infty}\left(1-\left(1-q^x\right)^T\right) d x,     
\end{align} 
where step $(a)$ is true since $\max_{\tau\in\{0,1,\ldots,T-1 \}} \Delta t_{\tau}$ is non-negative; $(b)$ follows the fact $\Delta t_\tau, \forall \tau \in \{0,1,\ldots,T-1\}$ are independently distributed; $(c)$ is true since $1-\left(1-q^x\right)^T$ decreases with respect to $x$.

By setting $u=1-q^x$, we have
\begin{align}
& \int_0^{\infty}\left(1-\left(1-q^x\right)^T\right) d x \nonumber\\
& =-\frac{1}{\log q} \int_0^1 \frac{1-u^T}{1-u} d u \nonumber\\
& =-\frac{1}{\log q} \int_0^1\left(1+u+\cdots+u^{T-1}\right) d u \nonumber \\
 & = -\frac{1}{\log q}\left(1+\frac{1}{2}+\cdots+\frac{1}{T}\right) \nonumber\\
& \leq -\frac{1}{\log q} \left( 1+\int_{1}^2\frac{1}{x}dx+\int_{2}^3\frac{1}{x}dx+\cdots+\int_{T-1}^{T}\frac{1}{x}dx \right) \nonumber\\
& = -\frac{1}{\log q} \left( 1+\int_{1}^{T}\frac{1}{x}dx \right) = -\frac{1+\log T}{\log q}. 
\end{align}
Therefore, we can prove 
$$
\bE\left[\max_{\tau\in\{0,1,\ldots,\tau' \}} \Delta t_{\tau} \right] \leq 1 - \frac{1+\log T}{(\log(1-\alpha(M)))}.
$$

\section{Proof of Proposition \ref{prop:regret}}
\label{App:proof:regret}

We rewrite the regret of the \lc{} algorithm as
\begin{align}
\text{Reg}(T)\triangleq& \sum_{t=0}^{T-1}\sum_{n=1}^{N}\left(\bE\left[\mu_nS_n^*(t)\right] -\bE\left[\mu_n\wh{S}_n(t)\right]\right)\nonumber\\
=&\sum_{t=0}^{T-1}\Delta R(t),
\end{align}
where $\Delta R(t)\triangleq \sum_{n=1}^{N}\bE\left[\mu_nS_n^*(t)-\mu_n\wh{S}_n(t)\right]$. 

Next, we consider the drift of the Lyapunov function $L(t)\triangleq\frac{1}{2}\sum_{n=1}^{N}Q_n^2(t)/\mu_n$. By adding the term $\eta\Delta R(t)$ on both sides of the drift, we obtain 
\begin{align*}
&\bE\left[L(t+1)-L(t)\right]+\eta \Delta R(t)\nonumber\\
&\stackrel{(a)}{\leq}\sum_{n=1}^{N}\bE\left[\frac{Q_n(t)(\lambda_n-\wh{S}_n(t)X_n(t)+\epsilon)}{\mu_n}\right]\nonumber\\
&+\eta\sum_{n=1}^{N}\bE\left[\mu_nS_n^*(t)\right]-\eta \sum_{n=1}^{N}\bE\left[\mu_n\wh{S}_n(t)\right]+\frac{N}{\mu_{\min}}\nonumber\\
&=\sum_{n=1}^{N}\bE\left[(Q_n(t)+ \eta \mu_n)(S_n^{*}(t)-\wh{S}_n(t))\right]\nonumber\\
&+\sum_{n=1}^{N}\bE\left[Q_n(t)\left(\frac{\lambda_n+\epsilon}{\mu_n}-S_n^*(t)\right)\right]+\frac{N}{\mu_{\min}}\nonumber\\
&\stackrel{(b)}{\leq} \frac{N}{\mu_{\min}} +\sum_{n=1}^{N}\bE\left[(Q_n(t)+\eta \mu_n)(S_n^{*}(t)-\wh{S}_n(t))\right],
\end{align*}
where step $(a)$ uses $\epsilon\leq\delta\leq 1$; $(b)$ follows from the fact that $\mb{S}^*$ is the stationary randomized policy that is independent of the current system state and $\bE[S_n^*(t)X_n(t)]\geq\lambda_n+\epsilon, \forall n$ holds for $\epsilon\leq\delta\leq 1$.

Dividing $\eta$ on both sides of the above inequality and summing over $t=0, 1,\ldots,T-1$, we have 
\begin{align}
\label{eqn:prop:reg:sumt}
& \text{Reg}(T) = \sum_{t=0}^{T-1}\Delta R(t) \nonumber\\
&\leq -\frac{1}{\eta}\bE\left[L(T)-L(0)\right] + \frac{NT}{\eta\mu_{\min}} \nonumber\\
&+\sum_{t=0}^{T-1}\frac{1}{\eta}\sum_{n=1}^{N}\bE\left[\left(Q_n(t)+\eta\mu_n\right)\left(S_n^*(t)-\wh{S}_n(t)\right)\right] \nonumber\\
&\leq \frac{NT}{\eta\mu_{\min}}+\frac{1}{\eta}\sum_{t=0}^{T-1}\sum_{n=1}^{N}\bE\left[\left(Q_n(t)+\eta\mu_n\right)\left(S_n^*(t)-\wh{S}_n(t)\right)\right],
\end{align}
where the last step uses the fact that $L(0)=0$ and $L(T)\geq0$.

Next, we focus on the term $$\sum_{n=1}^{N}\left(Q_n(t)+\eta\mu_n\right)\left(S_n^*(t)-\wh{S}_n(t)\right).$$ Then, we have
\begin{align}
\label{eqn:prop:reg:alg}
&\sum_{n=1}^{N}\left(Q_n(t)+ \eta\mu_n\right)\left(S_n^*(t)-\wh{S}_n(t)\right)\nonumber\\
\leq &\sum_{n=1}^{N}\left(Q_n(t)+\eta\mu_n\right)\left(\wt{S}_n(t)-\wh{S}_n(t)\right)\nonumber\\
=&\sum_{n=1}^{N}\left(Q_n(t)+\eta\mu_n\right)\wt{S}_n(t)-\sum_{n=1}^{N}\left(Q_n(t)+\eta w_n(t)\right)\wh{S}_n(t)\nonumber\\
 +& \eta\sum_{n=1}^{N} (w_{n}(t)-\mu_n)\wh{S}_n(t),
\end{align}
where the first inequality is true for $$\wt{\mb{S}}(t)\in\argmax_{\mb{S}\in\mc{S}}\sum_{n=1}^{N}\left(Q_n(t)+\eta\mu_n\right)S_n.$$

By substituting \eqref{eqn:prop:reg:alg} into \eqref{eqn:prop:reg:sumt}, we have
\begin{align}
\label{eqn:prop:reg:final}
&\text{Reg}(T) \leq \frac{NT}{\eta\mu_{\min}}+\underbrace{\sum_{t=0}^{T-1}\sum_{n=1}^{N}\bE\left[\left(w_n(t)-\mu_n\right)\wh{S}_n(t)\right]}_{\triangleq G_1(T)} \nonumber\\
& \left.
\begin{aligned}
& + \frac{1}{\eta}\sum_{t=0}^{T-1}\bE\Bigg[\sum_{n=1}^{N}\left(Q_n(t)+\eta\mu_n\right)\wt{S}_n(t) \nonumber \\
& \qquad\qquad\qquad-\sum_{n=1}^{N}\left(Q_n(t)+\eta w_n(t)\right)\wh{S}_n(t)\Bigg].   
\end{aligned}
\right\}\triangleq G_2(T) \\
\end{align}

Next, we focus on $G_1(T)$ and $G_2(T)$, respectively. Let $t_{n,l}$ denote the round at which arm $n$ was pulled, i.e., $\wh{S}_n(t_{n,l})=1$ and $\wh{S}_n(t_{n,l})=0$ if $t\neq t_{n,l}, l=1,2,\ldots, H_n(T)$. Therefore, we have $H_n(t_{n,l})=l-1$.

Let $G_{n,1}(T)\triangleq\sum_{t=0}^{T-1}\bE\left[\left(w_n(t)-\mu_n\right)\wh{S}_n(t)\right] $ and thus $G_1(T)=\sum_{n=1}^{N}G_{n,1}(T)$.

Hence, we have 
\begin{align}
\label{eqn:prop:reg:R1}
G_{n,1}(T)\stackrel{(a)}{\leq}&\sum_{t=0}^{T-1}\bE\left[(w_n(t)-\mu_n)\wh{S}_n(t)\id_{\mc{F}_n(t)}\right] \nonumber\\
\stackrel{(b)}{\leq}&\bE\left[\sum_{l=1}^{H_n(T)}(w_n(t_{n,l})-\mu_n)\id_{\mc{F}_n(t_{n,l})}\right] \nonumber\\
\stackrel{(c)}{\leq}&1+\bE\left[\sum_{l=2}^{H_n(T)}(w_n(t_{n,l})-\mu_n)\id_{\mc{F}_n(t_{n,l})}\right]\nonumber\\
\stackrel{(d)}{\leq}&1+\bE\left[\sum_{l=2}^{H_n(T)}(w_n(t_{n,l})-\mu_n)\id_{\mc{F}_n(t_{n,l})\cap\mc{G}_n(t_{n,l})}\right] \nonumber \\
& +\sum_{l=2}^{\infty}\bE\left[\id_{\ol{\mc{G}}_{n(t_{n,l})}}\right],
\end{align}
where step $(a)$ is true for $\mc{F}_n(t)\triangleq\{w_n(t)\geq\mu_n\}$ and $\id_{\{\cdot\}}$ being an indicator function; $(b)$ uses the definition of $t_{n,l}$, and the fact that $\wh{S}_n(t)\leq1,\forall t\geq0$; $(c)$ follows from the fact that $w_n(t)\leq1, \forall t\geq0$; $(d)$ is true for 
$$\mc{G}_n(t)\triangleq\left\{\ol{\mu}_n(t)-\mu_n\leq\sqrt{\frac{3\log t}{2H_n(t)}}\right\},$$
and $\ol{\mc{G}}_n(t)$ being the complement of the event $\mc{G}_n(t)$.

Next, we consider the second term on the right hand side (RHS) of \eqref{eqn:prop:reg:R1}. 
\begin{align}
\label{eqn:prop:reg:R1:second}
&\bE\left[\sum_{l=2}^{H_n(T)}(w_n(t_{n,l})-\mu_n)\id_{\mc{F}_n(t_{n,l})\cap\mc{G}_n(t_{n,l})}\right] \nonumber\\
\stackrel{(a)}{\leq}&\bE\left[\sum_{l=2}^{H_n(T)}2\sqrt{\frac{3\log t_{n,l}}{2H_n(t_{n,l})}}\right] \nonumber\\
\stackrel{(b)}{\leq}&\sqrt{6\log T}\bE\left[\sum_{l=2}^{H_n(T)}\frac{1}{\sqrt{l-1}}\right]\nonumber\\
\leq&\sqrt{6\log T}\left(1 + \int_{1}^{H_n(T)}\frac{1}{\sqrt{x}}dx\right)\nonumber\\
\leq&2\sqrt{6\log T}\bE\left[\sqrt{H_n(T)}\right],
\end{align}
where step $(a)$ uses the definition of $w_n(t)$ and $\mc{G}_n(t)$, and $(b)$ follows from the fact that $t_{n,l}\leq T$ and the definition of $t_{n,l}$. With regard to the third term on the RHS of \eqref{eqn:prop:reg:R1}, we have 
\begin{align*}
&\bE\left[\id_{\ol{\mc{G}}_n(t_{n,l})}\right]=\Pr\{\ol{\mc{G}}_n(t_{n,l})\}\nonumber\\
\stackrel{(a)}{\leq}&\Pr\left\{\bigcup_{m=l-1}^{T-1}\left\{\ol{\mu}_n(m)-\mu_n>\sqrt{\frac{3\log m}{2(l-1)}}\right\}\right\}\nonumber\\
\leq&\Pr\left\{\bigcup_{m=l-1}^{T-1}\left\{\ol{\mu}_n(m)-\mu_n>\sqrt{\frac{3\log m}{2m}}\right\}\right\}\nonumber\\
\stackrel{(b)}{\leq}&\sum_{m=l-1}^{T-1}\Pr\left\{\ol{\mu}_n(m)-\mu_n>\sqrt{\frac{3\log m}{2m}}\right\} \nonumber\\
\stackrel{(c)}{\leq}&\sum_{m=l-1}^{T-1}\frac{1}{m^3}\leq\frac{1}{(l-1)^3}+\int_{l-1}^{\infty}\frac{1}{x^3}dx\stackrel{(d)}{\leq}\frac{3}{2(l-1)^2},
\end{align*}
where step $(a)$ follows from the fact that 
$$\ol{\mc{G}}_n(t_{n,l})\subset\bigcup_{m=l-1}^{T-1}\left\{\ol{\mu}_n(m)-\mu_n>\sqrt{\frac{3\log m}{2(l-1)}}\right\};$$
$(b)$ uses the union bound; $(c)$ follows from the Chernoff-Hoeffding Bound (see, e.g., \cite[Fact 1]{auer2002finite}), i.e., for $X_1,X_2,\ldots,X_n$ be i.i.d. random variables with common range $[0,1]$ and mean $\mu$, then for any $a\geq0$, we have 
\begin{align}
\label{eqn:prop:reg:chernoff}
\Pr\left\{\frac{1}{n}\sum_{i=1}^{n}X_i\geq \mu+a\right\}\leq e^{-2na^2},
\end{align}
$(d)$ is true for $l\geq2$.

Hence, the third term on the RHS of \eqref{eqn:prop:reg:R1} can be bounded as follows.
\begin{align}
\label{eqn:prop:reg:R1:third}
\sum_{l=2}^{\infty}\bE\left[\id_{\ol{\mc{G}}_n(t_{n,l})}\right]\leq&\sum_{l=2}^{\infty}\frac{3}{2(l-1)^2}\leq \frac{\pi^2}{4},
\end{align}
where the last step use the fact that $\sum_{n=1}^{\infty}1/n^2=\pi^2/6$. By substituting \eqref{eqn:prop:reg:R1:second} and \eqref{eqn:prop:reg:R1:third} into \eqref{eqn:prop:reg:R1} and using the definition of $G_1(T)$, we have 
\begin{align}
\label{eqn:prop:reg:R1:final}
&G_1(T)\leq N\left(1+\frac{\pi^2}{4}\right) + 2\sqrt{6\log T}\sum_{n=1}^{N}\bE\left[\sqrt{H_n(T)}\right]\nonumber\\
\stackrel{(a)}{\leq}&N\left(1+\frac{\pi^2}{4}\right) + 2N\sqrt{6\log T}\bE\left[\sqrt{\frac{1}{N}\sum_{n=1}^{N}H_n(T)}\right] \nonumber\\
\stackrel{(b)}{\leq}&N\left(1+\frac{\pi^2}{4}\right) + 2\sqrt{6N S_{\max}T\log T},
\end{align}
where step $(a)$ uses Jensen's inequality; $(b)$ follows the fact that $\sum_{n=1}^{N}H_n(T)\leq TS_{\max}$.

Next, we consider the term $G_2(T)$ in \eqref{eqn:prop:reg:final}. $G_2(T)$ consists of the gap yielded in $t \in \mc{T}_0$ (i.e., $G_3(T)$) and the gap yielded in $t \notin \mc{T}_0$ (i.e., $G_4(T)$). Thus we have 
\begin{align}
\label{regret:G2}
&G_2(T) = \frac{1}{\eta}\sum_{t=0}^{T-1}\sum_{n=1}^{N}\bE\Bigg[\left(Q_n(t)+\eta\mu_n\right)\wt{S}_n(t) \nonumber \\
& \qquad\qquad\qquad\qquad\qquad\qquad-\left(Q_n(t)+\eta w_n(t)\right)\wh{S}_n(t)\Bigg]  \nonumber \\
&\left. 
\begin{aligned}
=& \frac{1}{\eta}\bE\Bigg[\sum_{t\in \mc{T}_0} \Bigg(\sum_{n=1}^{N}\left(Q_n(t)+\eta\mu_n\right)\wt{S}_n(t) \\ 
&\qquad\qquad -\sum_{n=1}^{N}\left(Q_n(t)+\eta w_n(t)\right)\wh{S}_n(t) \Bigg)\Bigg] \\
\end{aligned}
\right\}{\triangleq G_3(T)}  \nonumber \\
& \left.
\begin{aligned}
 &+ \frac{1}{\eta}\bE\Bigg[\sum_{t\notin \mc{T}_0}\Bigg(\sum_{n=1}^{N}\left(Q_n(t)+\eta\mu_n\right)\wt{S}_n(t) \\
 &\qquad\qquad\quad-\sum_{n=1}^{N}\left(Q_n(t)+\eta w_n(t)\right)\wh{S}_n(t) \Bigg)\Bigg].   
\end{aligned}
\right\}{\triangleq G_4(T)} 
\end{align}

Next, we consider $G_3(T)$ in \eqref{regret:G2}.
When $t=t_0=0$, we have 
\begin{align}
\label{prop:t0_gap}
&\sum_{n=1}^N \left(Q_n\left(0 \right)+\eta \mu_n\right) \wt{S}_n\left(0 \right)-\sum_{n=1}^N\left(Q_n\left(0 \right)+\eta w_n\left(0\right)\right) \wh{S}_n(0) \nonumber\\
\stackrel{(a)}{=} & \eta\sum_{n=1}^N  \mu_n \wt{S}_n\left(0\right)-\eta\sum_{n=1}^N w_n\left(0 \right) \wh{S}_n(0) \nonumber \\
\stackrel{(b)}{=}  & \eta\sum_{n=1}^N  \left(\mu_n-w_n(0)\right) \wt{S}_n (0),
\end{align}
where step $(a)$ is true due to $Q_n(0)=0,\forall n$; step $(b)$ follows from the fact that $|\mb{\wt{S}}(t)|=|\mb{\wh{S}}(t)|,\forall t$ and $w_n(0)=1,\forall n$.

When $\forall t \in \mc{T}_0\backslash\{t_0\}$, 
the gap between $\sum_{n=1}^N\left(Q_n(t)+\eta \mu_n\right) \wt{S}_n(t)$ and $\sum_{n=1}^N\left(Q_n(t)+\eta w_n(t)\right) \wh{S}_n(t)$ can be upper bounded as 
\begin{align}
\label{prop:t_tau_gap}
&\sum_{n=1}^N \left(Q_n\left(t\right)+\eta \mu_n\right) \wt{S}_n\left(t\right) \nonumber \\
& -\sum_{n=1}^N\left(Q_n\left(t\right)+\eta w_n\left(t\right)\right) \wh{S}_n(t) \nonumber\\
\stackrel{(a)}{=} & \sum_{n=1}^N \left(Q_n\left(t\right)+\eta \mu_n\right) \wt{S}_n\left(t\right) \nonumber \\
& -\sum_{n=1}^N\left(Q_n\left(t\right)+\eta w_n\left(t\right)\right) {S}^{\dagger}_n(t) \nonumber \\
\stackrel{(b)}{\leq} & \sum_{n=1}^N \left(Q_n\left(t\right)+\eta \mu_n\right)  \wt{S}_n\left(t\right) \nonumber\\
& -\sum_{n=1}^N\left(Q_n\left(t\right)+\eta w_n\left(t\right)\right) S^{\dagger}_n(t) \nonumber \\
& + \sum_{n=1}^N \left(Q_n\left(t\right)+\eta w_n(t)\right) S^{\dagger}_n\left(t\right) \nonumber \\
& -\sum_{n=1}^N\left(Q_n\left(t\right)+\eta w_n\left(t\right)\right) \wt{S}_n(t)   \nonumber\\
 = & \eta \sum_{n=1}^N \left( \mu_n  -  w_n\left(t\right)\right) \wt{S}_n(t),
\end{align}
where step $(a)$ uses the definition of $t \in \mc{T}_0$; $(b)$ uses the fact that $$\mb{S}^{\dagger}(t)\in\argmax_{\mb{S}\in\mc{S}}\sum_{n=1}^{N}\left(Q_n(t)+\eta w_n(t)\right)S_n.$$

Therefore, when $t \in \mc{T}_0$, the following inequality can always be satisfied
\begin{align}
\label{prop:t0_t_tau_gap}
&\sum_{n=1}^N \left(Q_n\left(t\right)+\eta \mu_n\right) \wt{S}_n\left(t\right)-\sum_{n=1}^N\left(Q_n\left(t\right)+\eta w_n\left(t\right)\right) \wh{S}_n(t) \nonumber\\
\leq & \eta\sum_{n=1}^N  \left(\mu_n-w_n(t)\right) \wt{S}_n (t).
\end{align}

Taking expectations and dividing $\eta$ on both sides to \eqref{prop:t0_t_tau_gap} and using the fact that $|\mc{T}_0|=\tau'\leq T-1$, we can have
\begin{align}
& G_3(T)  \leq \sum_{t=0}^{T-1}\sum_{n=1}^N\bE\left[ \left( \mu_n  -  w_n\left(t\right)\right) \wt{S}_n(t)\id_{\ol{\mc{F}}_n(t)} \right],
\end{align}
where $\ol{\mc{F}}_n(t)\triangleq\{w_n(t)<\mu_n\}$.

We recall the definition of $t_{n,l}$. Note that for $t\leq t_{n,1}$, we have $w_n(t)=1$ and thus $\mc{F}_n(t)$ happens. Therefore, we have 
\begin{align}
\label{prop:small_prob}
G_3(T) \leq &\sum_{n=1}^{N}\bE\left[\sum_{t=t_{n,1}+1}^{T-1}\left(\mu_n-w_n(t)\right)\wt{S}_n(t)\id_{\ol{\mc{F}}_n(t)}\right]\nonumber\\
\stackrel{(a)}{\leq}&\sum_{n=1}^{N}\bE\left[\sum_{t=t_{n,1}+1}^{T-1}\Pr\left\{\ol{\mu}_n(t)-\mu_n\leq-\sqrt{\frac{3\log t}{2H_n(t)}}\right\} \right]\nonumber\\
\leq&\sum_{n=1}^{N}\sum_{\nu=1}^{T-1}\sum_{m=1}^{\nu}\Pr\left\{\frac{1}{m}\sum_{i=1}^{m}X(i)-\mu_n\leq-\sqrt{\frac{3\log \nu}{2m}}\right\}\nonumber\\
\stackrel{(b)}{\leq}&\sum_{n=1}^{N}\sum_{\nu=1}^{T-1}\sum_{m=1}^{\nu}\frac{1}{\nu^3}=\sum_{n=1}^{N}\sum_{\nu=1}^{T-1}\frac{1}{\nu^2}\stackrel{(c)}{\leq}\frac{N\pi^2}{6},
\end{align}
where step $(a)$ follows from the fact that $\mu_n\leq 1$ and $\wt{S}_n(t)\leq 1$ as well as the definition of $\ol{\mc{F}}_n(t)$;
$(b)$ again uses the Chernoff-Hoeffding Bound (cf. \eqref{eqn:prop:reg:chernoff}); $(c)$ is true since $\sum_{\nu=1}^{T-1}1/\nu^2\leq\sum_{\nu=1}^{\infty}1/\nu^2=\pi^2/6$.

Next, we consider $G_4(T)$ in \eqref{regret:G2}. When  
$\forall t \in (t_\tau,t_{\tau+1}\wedge T)$ and $\tau \in \{0,1,\ldots,\tau'\}$, the gap between $\sum_{n=1}^N(Q_n(t)+\eta \mu_n) \wt{S}_n(t)$ and $\sum_{n=1}^N(Q_n(t_\tau)+\eta \mu_n)\wt{S}_n(t_{\tau})$ can be bounded as 

\begin{align}
\label{prop:opt_gap}
& \sum_{n=1}^N\left(Q_n(t)+\eta \mu_n\right) \wt{S}_n(t)  \nonumber\\
\stackrel{(a)}{\leq} &\sum_{n=1}^N\left(Q_n\left(t_\tau\right)+\eta \mu_n\right) \wt{S}_n(t) +(\Delta t_\tau-1) N \nonumber\\
\stackrel{(b)}{\leq}  &\sum_{n=1}^N\left(Q_n\left(t_\tau \right)+\eta \mu_n \right) \wt{S}_n\left(t_\tau\right)+  (\Delta t_\tau-1) N,
\end{align}
where step $(a)$ is true since $Q_{n}(t)\leq Q_{n}(t_\tau) + (\Delta t_\tau-1)  (\lambda_n +\epsilon), \forall n$, $\lambda_n +\epsilon \leq 1, \forall n$ and $\wt{S}_n(t)\leq 1,\forall n$; $(b)$ follows from the fact that $$\wt{\mb{S}}(t_{\tau})\in\argmax_{\mb{S}\in\mc{S}}\sum_{n=1}^{N}\left(Q_n(t_{\tau})+\eta\mu_n\right)S_n.$$


When $\forall t \in (t_\tau,t_{\tau+1}\wedge T)$ and $\tau \in \{0,1,\ldots,\tau'\}$, the gap between $\sum_{n=1}^N\left(Q_n(t)+\eta w_n(t)\right) \wh{S}_n(t)$ and $\sum_{n=1}^N\left(Q_n(t_\tau)+\eta w_n(t_\tau)\right) \wh{S}_n(t_{\tau})$ can be bounded as 
\begin{align}
\label{prop:our_policy_gap}
& \sum_{n=1}^N\left(Q_n(t)+\eta w_n(t)\right) \wh{S}_n(t) \nonumber \\
\stackrel{(a)}{\geq} &\sum_{n=1}^N\left(Q_n(t)+\eta w_n(t)\right) \wh{S}_n(t-1) \nonumber \\
\stackrel{(b)}{\geq} &\sum_{n=1}^N\big(Q_n(t-1)-1+\eta w_n(t-1) \nonumber \\
& +\eta\left(w_n(t)-w_n(t-1)\right)\big) \wh{S}_n(t-1) \nonumber \\
\stackrel{(c)}{\geq} &\sum_{n=1}^N\left(Q_n\left(t_\tau\right)+\eta w_n\left(t_\tau\right)\right) \wh{S}_n\left(t_\tau\right)  -\sum_{t^{\prime}=t_\tau+1}^t N \nonumber \\
& +\sum_{t^{\prime}=t_\tau+1}^t \sum_{n=1}^N \eta\left(w_n\left(t^{\prime}\right)-w_n\left(t^{\prime}-1\right)\right) \wh{S}_n\left(t^{\prime}-1\right) \nonumber \\
\stackrel{(d)}{\geq} &\sum_{n=1}^N\left(Q_n\left(t_\tau\right)+\eta w_n\left(t_\tau\right)\right) \wh{S}_n\left(t_\tau\right)  -\sum_{t^{\prime}=t_\tau+1}^{(t_{\tau+1}-1)\wedge(T-1)} N \nonumber \\
&+\sum_{t^{\prime}=t_\tau+1}^{(t_{\tau+1}-1)\wedge(T-1)} \sum_{n=1}^N \eta\left(w_n\left(t^{\prime}\right)-w_n\left(t^{\prime}-1\right)\right)^{-} \wh{S}_n\left(t^{\prime}-1\right) \nonumber \\
\stackrel{(e)}{\geq} & \sum_{n=1}^N\left(Q_n\left(t_\tau\right)+\eta w_n\left(t_\tau\right)\right)\wh{S}_n\left(t_\tau\right) -(\Delta t_{\tau}-1) N \nonumber \\
& -\sum_{t^{\prime}=t_\tau+1}^{(t_{\tau+1}-1)\wedge(T-1)} \sum_{n=1}^N \eta\Bigg(\frac{1}{H_n\left(t^{\prime}-1\right)}+\sqrt{\frac{3 \log T}{2 H_n\left((t^{\prime}-1)\right)}}\Bigg) \nonumber\\
& \qquad \cdot\id\left\{H_n(t'-1) \geq 1 \right\}\wh{S}_n\left(t^{\prime}-1\right) \nonumber\\
& - \eta\sum_{t^{\prime}=t_\tau+1}^{(t_{\tau+1}-1)\wedge (T-1)} \sum_{n=1}^N \id\left\{ 
\begin{aligned}
 &  H_n(t'-1)=0, \\
  &\wh{S}_n(t'-1)=1 
\end{aligned}
\right\},
\end{align}
where step $(a)$ uses the definition of the ``compare'' step of the algorithm; $(b)$ is true since $Q_n(t)\geq Q_n(t-1)-1,\forall n$ according to the update policy of the virtual-queue length; $(c)$ is from doing $(b)$ iteratively from $t$ to $t_{\tau}$; $(d)$ is true since $t< t_{\tau+1}$ and we denote $(x)^-\triangleq\min\{0,x\}$; $(e)$ uses lemma \ref{lemma:weight_change}.


After combining \eqref{prop:t0_t_tau_gap}, \eqref{prop:opt_gap}, \eqref{prop:our_policy_gap} and the definition of $\wh{\mb{S}}(t)$, we can have
\begin{align}
\label{prop:our_policy_opt_gap}
& \sum_{n=1}^N\left(Q_n(t)+\eta \mu_n\right) \wt{S}_n(t) -\sum_{n=1}^N\left(Q_n(t)+\eta w_n(t)\right) \wh{S}_n(t) \nonumber\\
\leq& \eta \sum_{n=1}^N \left( \mu_n  -  w_n\left(t_\tau\right)\right) \wt{S}_n(t_\tau)+2(\Delta t_{\tau}-1)N \nonumber \\
 & +\sum_{t^{\prime}=t_\tau+1}^{(t_{\tau+1}-1)\wedge (T-1)} \sum_{n=1}^N \eta\Bigg(\frac{1}{H_n\left(t^{\prime}-1\right)} +\sqrt{\frac{3 \log T}{2 H_n\left(t^{\prime}-1\right)}}\Bigg)\nonumber \\
 & \qquad\qquad\qquad\qquad\qquad \cdot\id\left\{H_n(t'-1) \geq 1 \right\}\wh{S}_n\left(t^{\prime}-1\right) \nonumber \\
 & + \eta\sum_{t^{\prime}=t_\tau+1}^{(t_{\tau+1}-1)\wedge (T-1)} \sum_{n=1}^N \id\left\{ 
 \begin{aligned}
 & H_n(t'-1)=0, \\
 & \wh{S}_n(t'-1)=1   
 \end{aligned}\right\}  .
\end{align}

Summing over $t \in (t_\tau,t_{\tau+1}\wedge T)$ in \eqref{prop:our_policy_opt_gap}, we can have 
\begin{align}
\label{prop:tau_frame_gap}
&\sum_{t=t_\tau+1}^{(t_{\tau+1}-1)\wedge(T-1)}
\Bigg(\sum_{n=1}^N\left(Q_n(t)+\eta \mu_n\right) \wt{S}_n(t)  \nonumber \\
& \qquad\qquad\qquad\qquad\qquad -\sum_{n=1}^N\left(Q_n(t)+\eta w_n(t)\right) \wh{S}_n(t) \Bigg)\nonumber \\
& \leq \eta (\Delta t_{\tau}-1)\sum_{n=1}^N \left( \mu_n  -  w_n\left(t_\tau\right)\right) \wt{S}_n(t_\tau)+2(\Delta t_{\tau}-1)^2 N  \nonumber \\
 & +\eta (\Delta t_{\tau}-1) \sum_{t'=t_\tau+1}^{(t_{\tau+1}-1)\wedge(T-1)}  \sum_{n=1}^N \Bigg(\frac{1}{H_n\left(t^{\prime}-1\right)}  \nonumber \\
 & +\sqrt{\frac{3 \log T}{2 H_n\left(t^{\prime}-1\right)}}\Bigg)  \id\left\{H_n(t'-1) \geq 1 \right\}\wh{S}_n\left(t^{\prime}-1\right) \nonumber \\
 &+ \eta (\Delta t_{\tau}-1)\sum_{t^{\prime}=t_\tau+1}^{(t_{\tau+1}-1)\wedge (T-1)} \sum_{n=1}^N \id\left\{ 
 \begin{aligned}
 & H_n(t'-1)=0, \\
 &\wh{S}_n(t'-1)=1   
 \end{aligned}\right\}.
\end{align}

Summing over $\tau \in \{0,1,\ldots,\tau'\}$ in \eqref{prop:tau_frame_gap}, the total cumulative gap over $t\notin \mc{T}_0$ is
\begin{align}
\label{prop:whole_frame_gap}
& \sum_{t \notin \mc{T}_0}\Bigg(\sum_{n=1}^N\left(Q_n(t)+\eta \mu_n\right) \wt{S}_n(t) \nonumber \\
& \qquad\qquad\qquad\qquad - \sum_{n=1}^N\left(Q_n(t)+\eta w_n(t)\right) \wh{S}_n(t)\Bigg) \nonumber \\
& \leq \sum_{\tau=0}^{\tau'}2(\Delta t_{\tau}-1)^2N \nonumber \\
& + \eta \sum_{\tau=0}^{\tau'} (\Delta t_{\tau}-1)\sum_{n=1}^N \left( \mu_n  -  w_n\left(t_\tau\right)\right) \wt{S}_n(t_\tau) \nonumber \\
& +\sum_{\tau=0}^{\tau'}\sum_{t^{\prime}=t_\tau+1}^{(t_{\tau+1}-1)\wedge (T-1)} \sum_{n=1}^N \eta(\Delta t_{\tau}-1)\Bigg(\frac{1}{H_n\left(t^{\prime}-1\right)} \nonumber \\
& +\sqrt{\frac{3 \log T}{2 H_n\left(t^{\prime}-1\right)}}\Bigg)  \id\left\{H_n(t'-1) \geq 1 \right\}\wh{S}_n\left(t^{\prime}-1\right) \nonumber \\
&+ \sum_{\tau=0}^{\tau'} \sum_{t^{\prime}=t_\tau+1}^{(t_{\tau+1}-1)\wedge (T-1)} \sum_{n=1}^N \eta (\Delta t_{\tau}-1) \id\left\{ 
\begin{aligned}
&H_n(t'-1)=0, \\
&\wh{S}_n(t'-1)=1 
\end{aligned}\right\}.
\end{align}

Taking expectations and dividing $\eta$ on both sides of \eqref{prop:whole_frame_gap}, we can derive the upper bound of term $G_4(T)$ as 
\begin{align}
\label{prop:G4}
& G_4(T) \stackrel{(a)}{\leq}\frac{2C_1(M)NT}{\eta} \nonumber \\
& +\bE\left[\sum_{\tau=0}^{\tau'} (\Delta t_{\tau}-1)\sum_{n=1}^N \left( \mu_n  -  w_n\left(t_\tau\right)\right) \wt{S}_n(t_\tau)\right] \nonumber \\
& + \bE\Bigg[\sum_{\tau=0}^{\tau'}\sum_{t^{\prime}=t_\tau+1}^{(t_{\tau+1}-1)\wedge (T-1)} \sum_{n=1}^N (\Delta t_{\tau}-1)\Bigg(\frac{1}{H_n\left(t^{\prime}-1\right)} \nonumber \\
&  +\sqrt{\frac{3 \log T}{2 H_n\left(t^{\prime}-1\right)}}\Bigg)\id\left\{H_n(t'-1) \geq 1 \right\}\wh{S}_n\left(t^{\prime}-1\right)\Bigg]\nonumber\\
& +\bE\Bigg[\sum_{\tau=0}^{\tau'} \sum_{t^{\prime}=t_\tau+1}^{(t_{\tau+1}-1)\wedge (T-1)} \sum_{n=1}^N (\Delta t_{\tau}-1) \nonumber \\
& \qquad\qquad \cdot \id\left\{ H_n(t'-1)=0, \wh{S}_n(t'-1)=1\right\}\Bigg] \nonumber\\
&\stackrel{(b)}{\leq} \frac{2C_1(M) NT}{\eta}+ C_2(M) N  \nonumber \\
& +\bE\left[\sum_{\tau=0}^{\tau'} (\Delta t_{\tau}-1)\sum_{n=1}^N \left( \mu_n  -  w_n\left(t_\tau\right)\right) \wt{S}_n(t_\tau)\right] \nonumber \\
& + \bE\Bigg[\sum_{\tau=0}^{\tau'}\sum_{t^{\prime}=t_\tau+1}^{(t_{\tau+1}-1)\wedge (T-1)} \sum_{n=1}^N (\Delta t_{\tau}-1)\Bigg(\frac{1}{H_n\left(t^{\prime}-1\right)} \nonumber \\
&  +\sqrt{\frac{3 \log T}{2 H_n\left(t^{\prime}-1\right)}}\Bigg)\id\left\{H_n(t'-1) \geq 1 \right\}\wh{S}_n\left(t^{\prime}-1\right)\Bigg],
\end{align}
where step $(a)$ is due to $\tau'\leq T-1$ and $C_1(M)\triangleq(\alpha^2(M)-3\alpha(M)+2)/\alpha^2(M)=\bE\left[(\Delta t_{\tau}-1)^2\right]$ due to Lemma \ref{lemma:exp_time_gap}; $(b)$ uses the fact that $\id\{H_n(t)=0, \wh{S}_n(t)=1\}=1$ only happens once over $T$ rounds and $C_2(M)\triangleq 1/\alpha(M) -1 = \bE\left[\Delta t_{\tau}-1\right]$ due to Lemma \ref{lemma:exp_time_gap}.

The third term in \eqref{prop:G4} can be bounded as
\begin{align}
& \sum_{n=1}^N \bE\left[\sum_{\tau=0}^{\tau'}(\Delta t_{\tau}-1)\left( \mu_n  -  w_n\left(t_\tau\right)\right)^{+}\right] \nonumber \\
& \stackrel{(a)}{\leq} \sum_{n=1}^N\sum_{\tau=0}^{T-1} \bE\left[ (\Delta t_{\tau} -1) \left( \mu_n  -  w_n\left(t_\tau\right)\right)^{+} \right]\nonumber \\ 
& \stackrel{(b)}{\leq} C_2(M) \sum_{n=1}^N\sum_{t=0}^{T-1} \bE\left[  \left( \mu_n  -  w_n\left(t\right)\right)^{+} \right] \nonumber\\
& \stackrel{(c)}{\leq} C_2(M) \frac{N \pi^2}{6},
\end{align}
where step $(a)$ follows from the fact that $\tau'\leq T-1$; $(b)$ is true since $C_2(M)\triangleq 1/\alpha(M)-1$ due to Lemma \ref{lemma:exp_time_gap}; $(c)$ is true due to \eqref{prop:small_prob}.

The fourth term in \eqref{prop:G4} can be bounded as
\begin{align}
& \sum_{n=1}^N\bE\Bigg[\sum_{\tau=0}^{\tau'}\sum_{t^{\prime}=t_\tau+1}^{{t_{\tau+1}-1}\wedge(T-1)}(\Delta t_{\tau}-1)\Bigg(\frac{1}{H_n\left(t^{\prime}-1\right)} \nonumber \\
& +\sqrt{\frac{3 \log T}{2 H_n\left(t^{\prime}-1\right)}}\Bigg) \id\left\{H_n(t'-1) \geq 1 \right\}\wh{S}_n\left(t^{\prime}-1\right)  \Bigg] \nonumber\\
& \leq  \sum_{n=1}^N \bE\Bigg[ \left(\max_{\tau \in \{0,1,\ldots,\tau'\}} (\Delta t_{\tau}-1)\right)\cdot\nonumber \\
&  \sum_{\tau = 0}^{\tau'} \sum_{t'=t_\tau+1}^{(t_{\tau+1}-1) \wedge (T-1)}\frac{\id\left\{H_n(t'-1) \geq 1 \right\}\wh{S}_n\left(t^{\prime}-1\right)}{H_n\left(t^{\prime}-1\right)}
\Bigg]  \nonumber \\
& + \sqrt{\frac{3\log T}{2}}\sum_{n=1}^N \bE\Bigg[ \left(\max_{\tau \in \{0,1,\ldots,\tau'\}} (\Delta t_{\tau}-1)\right)\cdot\nonumber \\
& \sum_{\tau = 0}^{\tau'} \sum_{t'=t_\tau+1}^{(t_{\tau+1}-1) \wedge (T-1)}\frac{\id\left\{H_n(t'-1) \geq 1 \right\}\wh{S}_n\left(t^{\prime}-1\right)}{\sqrt{H_n\left(t^{\prime}-1\right)}}\Bigg]  \nonumber \\
& \stackrel{(a)}{\leq}  \sum_{n=1}^N \bE\left[ \left(\max_{\tau \in \{0,1,\ldots,\tau'\}} (\Delta t_{\tau}-1)\right)\sum_{l = 2}^{H_n(T)} \frac{1}{l-1} \right] \nonumber \\
& + \sqrt{\frac{3\log T}{2}} \sum_{n=1}^N \bE\left[ \left(\max_{\tau \in \{0,1,\ldots,\tau'\}} (\Delta t_{\tau}-1)\right)\sum_{l = 2}^{H_n(T)} \frac{1}{\sqrt{l-1}}\right] \nonumber\\
& \leq  \sum_{n=1}^N \bE\left[ \left(\max_{\tau \in \{0,1,\ldots,\tau'\}} (\Delta t_{\tau}-1)\right)\left(1+\int_{1}^{H_n(T)} \frac{1}{x}dx \right) \right] \nonumber \\
& + \sqrt{\frac{3\log T}{2}} \sum_{n=1}^N \bE\Bigg[ \left(\max_{\tau \in \{0,1,\ldots,\tau'\}} (\Delta t_{\tau}-1)\right)\cdot \nonumber \\
& \qquad\qquad\qquad \qquad\qquad\qquad \left(1+\int_{1}^{H_n(T)} \frac{1}{\sqrt{x}}dx \right)\Bigg]  \nonumber \\
& \leq  \sum_{n=1}^N \bE\left[ \left(\max_{\tau \in \{0,1,\ldots,\tau'\}} (\Delta t_{\tau}-1)\right)\log H_n(T) \right] \nonumber \\
& + \sqrt{\frac{3\log T}{2}} \sum_{n=1}^N \bE\left[ \left(\max_{\tau \in \{0,1,\ldots,\tau'\}} (\Delta t_{\tau}-1)\right)\sqrt{H_n(T)}\right] \nonumber \\
& \stackrel{(b)}{\leq} \frac{1+\log T}{C_3(M)}N\log \frac{S_{\max}T}{N}  + \frac{\left(1+\log T\right)\sqrt{6NS_{\max}T\log T}}{2C_3(M)},
\end{align}
where step $(a)$ is similar to step $(b)$ in \eqref{eqn:prop:reg:R1:second}; $(b)$ uses lemma \ref{lemma:exp_time_gap}, the convexity of $\log x$ and $\sqrt{x}$, and $C_3(M) \triangleq -\log (1-\alpha(M))$.

Thus,
$G_4(T)$ can be bounded as 
\begin{align}
    & G_4(T) \leq  \frac{2C_1(M)NT}{\eta} +C_2(M) N+C_2(M)\frac{N\pi^2}{6} \nonumber \\
    & + \frac{1+\log T}{C_3(M)}N\log \frac{S_{\max}T}{N}  + \frac{\left(1+\log T\right)\sqrt{6NS_{\max}T\log T}}{2C_3(M)}.
\end{align}

In conclusion,
\begin{align}
&\text{Reg}(T)\leq  \frac{NT}{\eta\mu_{\min}}  +2\sqrt{6NS_{\max}T\log T} +N\left(1+\frac{5\pi^2}{12}\right)  \nonumber\\
& + \frac{2C_1(M)NT}{\eta} + C_2(M) N+C_2(M)\frac{N\pi^2}{6}  + \frac{1+\log T}{C_3(M)}  \nonumber \\
& \cdot N\log \frac{S_{\max}T}{N} + \frac{\left(1+\log T\right)\sqrt{6NS_{\max}T\log T}}{2C_3(M)}.  
\end{align}

\section{Proof of Lemma \ref{lemma:large_prob_weight_gap}}
\label{app_lemma:large_prob_weight_gap}
Since the LCFL algorithm independently picks subsets at uniformly random in each $t$, there exists a $D>0$ such that
\begin{align*}
& \Pr\left(\mb{S}^{\ddagger}(t)\in \mc{R}(k) \text{ for some } k \in \{t-D,...,t\}\right)  \nonumber \\
& \geq \gamma(M,D),
\end{align*}
where $\gamma(M,D)\triangleq 1-(1-M/|\mc{S}|)^D$ and $D$ is selected subject to $\gamma(M,D)\geq 1/(1+\delta)$. 

Under the LCFL algorithm, we have 
\begin{align}
\label{prop:exp_gap_loose_2}
& \sum_{n=1}^N\left(Q_n(t)+\eta w_n(t)\right) \wh{S}_n(t)  \nonumber \\
\stackrel{(a)}{\geq} &\sum_{n=1}^N\left(Q_n(t)+\eta w_n(t)\right) \wh{S}_n(t-1)\nonumber \\
\stackrel{(b)}{\geq} &\sum_{n=1}^N\left(Q_n(t-1)-1+\eta w_n(t-1)-\eta\right) \wh{S}_n(t-1)   \nonumber \\
\stackrel{(c)}{\geq} &\sum_{n=1}^N\left(Q_n(t-1)+\eta w_n(t-1)\right) \wh{S}_n(t-1)-(1+\eta)N   \nonumber \\
\stackrel{(d)}{\geq} & \sum_{n=1}^N\left(Q_n\left(k\right)+\eta w_n\left(k\right)\right) \wh{S}_n\left(k\right)  -(t-k)N(1+\eta)  \nonumber \\
\stackrel{(e)}{\geq} & \sum_{n=1}^N\left(Q_n\left(k\right)+\eta w_n\left(k\right)\right) \wh{S}_n\left(k\right) -ND(1+\eta) ,
\end{align}
where step $(a)$ follows the ``compare'' step of the LCFL algorithm; $(b)$ uses the fact that the virtual queue length can at most reduce by one in each round and the definition of UCB weight, i.e., $Q_n(t)\geq Q_n(t-1)-1$ and $w_n(t)\geq w_n(t-1)-1, \forall n$; $(c)$ uses the fact that $\wh{S}_n(t)\leq1, \forall t$; $(d)$ follows by iterating $(c)$ until round $k$; $(e)$ is true since $k\geq t-D$.

When event $\mb{S}^{\ddagger}(t) \in \mc{R}(k)$ happens, 
based on \eqref{prop:exp_gap_loose_2}, we have 
\begin{align}
 & \sum_{n=1}^N\left(Q_n(t)+\eta w_n(t)\right) \wh{S}_n(t)  \nonumber \\
\stackrel{(a)}{\geq} & \sum_{n=1}^N\left(Q_n(k)+\eta w_n(k)\right) R_n(k)  -ND(1+\eta) \nonumber \\
\stackrel{(b)}{\geq} & \sum_{n=1}^N\left(Q_n(k)+\eta w_n(k)\right) S^{\ddagger}_n(t) -ND(1+\eta) \nonumber \\
\stackrel{(c)}{\geq} & \sum_{n=1}^N Q_n(t)S^{\ddagger}_n(t) - ND(2+\eta),
\end{align}
where step $(a)$ is true due to the ``compare'' step in our \lc{} algorithm where $\mb{R}(k)\in \argmax_{\mb{R}\in
\mc{R}(k)}\sum_{n=1}^{N} \left(Q_n(k)+\eta w_n(k)\right)R_n$; $(b)$ is true when event $\mb{S}^{\ddagger}(t) \in \mc{R}(k)$ happens; $(c)$ is true since $Q_n (t)\leq Q_n(k) + D$ and $S^{\ddagger}_n(t)\leq 1,\forall n$.

This implies that when event $\mb{S}^{\ddagger}(t) \in \mc{R}(k)$ happens, we have 
\begin{align}
\label{eqn:prop:vol:V1:TLFG}
 \sum_{n=1}^NQ_n(t)\wh{S}_{n}(t)& \geq\sum_{n=1}^NQ_n(t)S^{\ddagger}_{n}(t)-N\eta-ND(2+\eta) \nonumber\\
 & \geq \sum_{n=1}^NQ_n(t)S^{\ddagger}_{n}(t)-B_1,
\end{align}
where the first step uses the fact that $w_{n}(t)\leq 1, \forall n$, and the second step is true for $B_1\triangleq 2ND(1+\eta)$ and uses the fact that $D\geq1$. 
Noting that the probability event $\mb{S}^{\ddagger}(t) \in \mc{R}(k)$ happens is at least $\gamma(M,D)$, we have the desired result.

\section{Proof of Proposition \ref{prop:zero_violation}}
\label{App:proof:zero_violation}

Select the Lyapunov function
\begin{align}
V(t) \triangleq \|\mb{W}(t)\|_2,
\end{align}
where $\mb{W}(t)\triangleq (\mb{Q}(t)/\sqrt{\bs{\mu}})$ and we have the following key lemma that characterizes the conditional expected drift given the current state $\mb{I}(t)\triangleq(\mb{Q}(t),\mb{w}(t))$ when $V(t)$ is large enough, which is shown in Appendix \ref{APP:lemma:drift}.
\begin{lemma}
\label{lemma:drift}
For any $\epsilon\leq (\gamma\delta+\gamma-1)/2 $, if $V\geq U(\eta,\gamma)\triangleq 2N/(\mu_{\min}(\gamma\delta+\gamma-1))+8 ND(1+\eta)/(\gamma\delta+\gamma-1)$, 
then 
\begin{align}
\bE\left[V(t+1)-V(t)\middle|\mb{I}(t)\right]\leq -\frac{\gamma\delta+\gamma-1}{4}.
\end{align}
Moreover, 
\begin{align*}
\left|V(t+1)-V(t)\right|
\leq \frac{N}{\mu_{\min}}\triangleq \zeta,
\end{align*}
where we recall that $\mu_{\min}\triangleq\min_{n}\mu_{n}>0$ .
\end{lemma}

For any $\epsilon\leq (\gamma\delta+\gamma-1)/2 $, Lemma \ref{lemma:drift} satisfies the conditions of \cite[Lemma 11]{liu2021efficient} and thus we have 
\begin{align}
\bE\left[e^{\theta(\gamma) V(t)}\right]& \leq e^{\theta(\gamma)  V(0)} \nonumber\\
&+v_0(\gamma)e^{\theta(\gamma)(\zeta+U(\eta,\gamma))},     
\end{align}
where $\theta(\gamma)\triangleq3(\gamma\delta+\gamma-1) /(12\zeta^2+ (\gamma\delta+\gamma-1)\zeta)$ and $v_0(\gamma)\triangleq 8/((\gamma\delta+\gamma-1)\theta(\gamma))$.

According to Jensen's inequality for convex function $e^{\theta x}$, we have 
\begin{align*}
e^{\theta(\gamma)\bE[V(t)]}& \leq\bE\left[e^{\theta(\gamma) V(t)}\right] \nonumber \\
&\leq e^{\theta(\gamma)  V(0)}+v_0(\gamma)e^{\theta(\gamma)(\zeta+ U(\eta,\gamma))},
\end{align*}
which implies 
\begin{align}
&\bE[V(t)] \nonumber \\
\leq & \frac{1}{\theta(\gamma)}\log\left(e^{\theta(\gamma)  V(0)}+v_0(\gamma) e^{\theta(\gamma)(\zeta+ U(\eta,\gamma))}\right)\nonumber\\
\stackrel{(a)}{\leq}&\frac{1}{\theta(\gamma)}\log\left((v_0(\gamma)+1)e^{\theta(\gamma)(V(0)+\zeta+ U(\eta,\gamma))}\right)\nonumber\\
\stackrel{(b)}{=}&\frac{1}{\theta(\gamma)}\log (v_0(\gamma)+1) +\zeta+ U(\eta,\gamma),
\end{align}
where step $(a)$ is true since $v_0(\gamma) \geq1$; $(b)$ is true for $V(0)=0$.

According to the dynamics of virtual queues (cf. \eqref{eqn:virtualQ}), we have 
\begin{align*}
Q_n(t+1)\geq Q_n(t)+\lambda_n-\wh{S}_n(t)X_n(t)+\epsilon.     
\end{align*}
By summing the above inequality over $s=0,1,\ldots,t-1$ and taking expectation on both sides, we have
\begin{align}
\bE[Q_n(t)]\geq\sum_{s=0}^{t-1}(\lambda_n-\bE[\wh{S}_n(s)X_n(s)])+\epsilon t.
\end{align}
Using the fact that $V(t)=\|\mb{Q}(t)/\mb{\sqrt{\bs{\mu}}}\|_2\geq\|\mb{Q}(t)\|_1/\sqrt{N}$ due to $\mu_n\leq1, \forall n$, we have 
\begin{align}
\label{eqn:prop:vlo:queue_length}
\bE\left[\|\mb{Q}(t)\|_1\right]\leq\sqrt{N}\left(\frac{1}{\theta(\gamma)}\log (v_0(\gamma)+1)+\zeta+U(\eta,\gamma)\right). 
\end{align}
As such, we have 
\begin{align}
\label{eqn:prop:vol:main}
&\left(\sum_{s=0}^{t-1}(\lambda_n-\bE[\wh{S}_n(s)X_n(s)])\right)^{+}\leq\left(\bE[Q_n(t)]-\epsilon t\right)^{+}\nonumber\\
\leq&\left(\bE[\|\mb{Q}(t)\|_1]- \epsilon t\right)^{+}\nonumber\\
\leq & \left( g_0(\eta)-\epsilon t\right)^{+} .
\end{align}
where $g_0(\eta)\triangleq\sqrt{N}\bigg(\frac{1}{\theta(\gamma)}\log (v_0(\gamma)+1)+\zeta+U(\eta,\gamma) \bigg)$.

\section{Proof of Lemma \ref{lemma:drift}}
\label{APP:lemma:drift}
In the rest of the proof, we omit the round index $t$ properly without causing confusion and use $Y^{+}$ to denote $Y(t+1)$, where $\mb{I}\triangleq(\mb{Q},\mb{w})$.
\begin{align}
\label{eqn:prop:main}
\Delta V\triangleq&\bE\left[V^+-V\middle|\mb{I}\right]\nonumber\\
=&\bE\left[\sqrt{\|\mb{W}^+\|_2^2}-\sqrt{\|\mb{W}\|_2^2}\middle|\mb{I}\right]\nonumber\\
\leq&\frac{1}{2\|\mb{W}\|_2}\bE\left[\|\mb{W}^+\|_2^2-\|\mb{W}\|_2^2\middle|\mb{I}\right]  \nonumber\\
=&\frac{\Delta V_1}{2\|\mb{W}\|_2},
\end{align}
where the second last step follows from the fact that $f(x)=\sqrt{x}$ is concave for $x>0$ and thus $f(x_1)-f(x_2)\leq f'(x_2)(x_1-x_2)=(x_1-x_2)/(2\sqrt{x_2})$ with $x_1=\|\mb{W}^+\|^2_2$ and $x_2=\|\mb{W}\|^2_2$, and the last step is true for $\Delta V_1=\bE\left[\|\mb{W}^+\|_2^2-\|\mb{W}\|_2^2\middle|\mb{I}\right]$.

Next, we consider the term $\Delta V_1$.
\begin{align}
\label{eqn:prop:vol:V1}
&\Delta V_1=\sum_{n=1}^N\bE\left[\frac{(Q_n^+)^2}{\mu_n}-\frac{(Q_n)^2}{\mu_n} \middle|\mb{I}\right]\nonumber\\
\stackrel{(a)}{\leq}&\sum_{n=1}^N\bE\left[\frac{(Q_n+\lambda_n-\wh{S}_nX_n+\epsilon)^2}{\mu_n}-\frac{Q_n^2}{\mu_n} 
 \middle|\mb{I}\right]\nonumber\\
\stackrel{(b)}{\leq}&2\sum_{n=1}^N\frac{1}{\mu_n}(\lambda_n+\epsilon)Q_n-2\sum_{n=1}^N\bE\left[Q_n\wh{S}_{n}\middle|\mb{I}\right]+\frac{N}{\mu_{\min}},
\end{align}
where step $(a)$ follows from the fact that $(\max\{x,0\})^2\leq x^2$ for any real number $x$; $(b)$ uses the fact that $\lambda_n+\epsilon\leq \mu_n\leq 1$.

Applying lemma \ref{lemma:large_prob_weight_gap} to \eqref{eqn:prop:vol:V1}, we have
\begin{align}
\label{eqn:prop:vol:V1:medium}
 &\Delta V_1  \leq  2\sum_{n=1}^N\frac{1}{\mu_n}(\lambda_n+\epsilon)Q_n + \frac{N}{\mu_{\min}}-2\sum_{n=1}^N\bE\left[Q_n\wh{S}_{n}\middle|\mb{I}\right] \nonumber \\
& \stackrel{(a)}{\leq}  2\sum_{n=1}^N\frac{1}{\mu_n}(\lambda_n+\epsilon)Q_n + \frac{N}{\mu_{\min}}  \nonumber \\
&\qquad\qquad\qquad -2\sum_{n=1}^N\bE\left[Q_n\wh{S}_{n}\middle|\mb{I},\mc{J}\right]\Pr\left(\mc{J}\right) \nonumber \\
& \stackrel{(b)}{\leq}  2\sum_{n=1}^N\frac{1}{\mu_n}(\lambda_n+\epsilon)Q_n + \frac{N}{\mu_{\min}} \nonumber \\
&\qquad \qquad \qquad-2\sum_{n=1}^N \left(Q_n{S}^{\ddagger}_{n}-B_1 \right) \Pr\left(\mc{J}\right) \nonumber \\
&  \stackrel{(c)}{\leq}  2\sum_{n=1}^N\frac{1}{\mu_n}(\lambda_n+\epsilon)Q_n + \frac{N}{\mu_{\min}}  -2 \gamma\sum_{n=1}^{N}Q_n{S}^{\ddagger}_n \nonumber \\
& \qquad\qquad\qquad\qquad\qquad\qquad +2B_1 \Pr\left(\mc{J}\right)  \nonumber \\
 & \stackrel{(d)}{\leq}  2\sum_{n=1}^N\frac{1}{\mu_n}(\lambda_n+\epsilon)Q_n + \frac{N}{\mu_{\min}}  -2 \gamma\sum_{n=1}^{N}Q_n{S}^{\ddagger}_n +2B_1,
\end{align}
where step $(a)$ is true since 
$\mc{J}\triangleq \left\{\sum_{n=1}^{N}Q_n{S}^{\ddagger}_n-\sum_{n=1}^{N}Q_n\wh{S}_n \leq B_1\right\}$; $(b)$ uses the fact that given state $\mb{I}$,
$\mb{S}^{\ddagger}$ is deterministic; $(c)$ follows from lemma \ref{lemma:large_prob_weight_gap}; $(d)$ is true since $\Pr\left(\mc{J}\right)\leq 1$.

Note that there exists non-negative numbers $\beta(\mb{s})$ with $\sum_{\mb{s}\in\mc{S}}\beta(\mb{s})=1$ satisfying
\begin{align*}
\lambda_n+\delta\leq\sum_{\mb{s}\in\mc{S}}\beta(\mb{s})s_{n}\mu_{n}, \forall n.
\end{align*}
Hence, we have 
\begin{align}
\label{eqn:prop:vol:V1:ranpolicy}
\sum_{n=1}^N\frac{(\lambda_n+\delta)Q_n}{\mu_n}\leq&\sum_{\mb{s}\in\mc{S}}\beta(\mb{s})\sum_{n=1}^NQ_{n}s_{n}\nonumber\\
\leq&\sum_{\mb{s}\in\mc{S}}\beta(\mb{s})\sum_{n=1}^NQ_nS_{n}^{\ddagger}\nonumber\\
=&\sum_{n=1}^NQ_nS_{n}^{\ddagger}.
\end{align}
Substituting \eqref{eqn:prop:vol:V1:ranpolicy} into \eqref{eqn:prop:vol:V1:medium}, we have 
\begin{align}
\label{eqn:prop:vol:V1:final}
\Delta V_1 & \stackrel{(a)}{\leq} -2(\gamma\delta-\epsilon+\gamma -1)\sum_{n=1}^{N}\frac{Q_n}{\mu_n}+ \frac{N}{\mu_{\min}}+2 B_1 \nonumber\\
& \stackrel{(b)}{\leq}-2(\gamma\delta-\epsilon+\gamma -1)\|\mb{W}\|  + B_2,
\end{align}
where step $(a)$ is true since we use the fact that $\lambda_n \leq 1,\forall n$; $(b)$ follows the fact that  $\|\mb{W}\|_1\geq\|\mb{W}\|$, $\sqrt{\mu_n} \geq \mu_n, \forall n$ and $B_2 \triangleq  N/\mu_{\min}+2B_1 =N/\mu_{\min} +4 ND(1+\eta)$.

Substituting \eqref{eqn:prop:vol:V1:final} into \eqref{eqn:prop:main}, we have 
\begin{align*}
&\Delta V \leq\frac{1}{2\|\mb{W}\|}\left(-2(\gamma\delta-\epsilon+\gamma -1 )\|\mb{W}\|+ B_2 \right)\nonumber\\
&=-(\gamma\delta-\epsilon+\gamma -1)+\frac{B_2}{2V}. 
\end{align*}
This implies that for any $\epsilon\leq (\gamma\delta+\gamma-1)/2$, if $V\geq U(\eta,\gamma)\triangleq 2N/(\mu_{\min}(\gamma\delta+\gamma-1))+8 ND(1+\eta)/(\gamma\delta+\gamma-1)$, then $\Delta V\leq-(\gamma\delta+\gamma-1)/4$.

In addition, 
\begin{align}
&\left|V^{+}-V\right|\nonumber\\
=&\left|\|\frac{\mb{Q}^+}{\sqrt{\bs{\mu}}}\|-\|\frac{\mb{Q}}{\sqrt{\bs{\mu}}} \|\right|\nonumber\\
\stackrel{(a)}{\leq}&\|\frac{\mb{Q}^+-\mb{Q}}{\sqrt{\bs{\mu}}}\| \nonumber\\
\stackrel{(b)}{\leq}&\| \frac{\mb{Q}^+-\mb{Q}}{\sqrt{\bs{\mu}}}  \|_1 \nonumber\\
\leq& N\max_n\|\frac{Q_n^+-Q_n}{\sqrt{\mu_{n}}}\|\stackrel{(c)}{\leq} \frac{N}{\mu_{\min}},
\end{align}
where step $(a)$ uses the fact that $|\|\mb{x}\|-\|\mb{y}\||\leq \|\mb{x}-\mb{y}\|$ for vectors $\mb{x}$ and $\mb{y}$; $(b)$ is true since $\|\mb{x}\|\leq\|\mb{x}\|_1$; $(c)$ is true since $\lambda_n\leq1$, $\wh{S}_{n}\leq1$, and $\epsilon\leq1$.

\bibliographystyle{IEEEtran}
\bibliography{refs}

\end{document}